\documentclass[lettersize,journal]{IEEEtran}
\usepackage{amsmath,amsfonts}
\usepackage{algorithmic}
\usepackage{algorithm}
\usepackage{array}
\usepackage{textcomp}
\usepackage{stfloats}
\usepackage{subfigure}
\usepackage{url}
\usepackage{verbatim}
\usepackage{graphicx}
\usepackage{cite}
\usepackage{xcolor}
\usepackage{hyperref}
\usepackage{tabularx}
\usepackage{caption}
\usepackage{booktabs}
\usepackage{xspace}
\usepackage{makecell}
\usepackage{algorithm}
\usepackage{algorithmic}
\usepackage{colortbl}
\usepackage{adjustbox}
\definecolor{Gray}{gray}{0.92}
\definecolor{sclgreyblue}{rgb}{0.2,0.3,0.5}%
\usepackage{mathtools}
\usepackage{amsthm}
\usepackage{multirow}
\usepackage{array} 
\usepackage{amsmath,bm,bbm}
\usepackage{mathtools}
\usepackage{bbding}
\usepackage{pifont}
\usepackage{wasysym}
\usepackage{wrapfig}
\usepackage{enumitem}
\newcommand{\abbr}[0]{LogoRA\xspace}

\newcolumntype{g}{>{\columncolor{Gray}}c}
\begin{document}

\title{LogoRA: Local-Global Representation Alignment for Robust Time Series Classification}

\author{Huanyu Zhang, 
        Yi-Fan Zhang$^\dagger$, 
        Zhang Zhang$^\dagger$, \IEEEmembership{Member, IEEE}, \\
        Qingsong Wen, \IEEEmembership{Senior Member, IEEE},
        and Liang Wang, \IEEEmembership{Fellow, IEEE}
\thanks{$^{\dagger}$~Corresponding Author}
\thanks{Huanyu Zhang, Yi-Fan Zhang, Zhang Zhang, and Liang Wang are with the State Key Laboratory of Multimodal Artificial Intelligence Systems (MAIS),  Institute of Automation, Chinese Academy of Sciences (CASIA), Beijing 100190, China, and also with the School of Artificial Intelligence, University of Chinese Academy of Sciences (UCAS), Beijing 100049, China.}
\thanks{Qingsong Wen is with the Squirrel AI Group.}
}

\markboth{Journal of \LaTeX\ Class Files,~Vol.~14, No.~8, August~2021}%
{Shell \MakeLowercase{\textit{et al.}}: A Sample Article Using IEEEtran.cls for IEEE Journals}


\maketitle

\begin{abstract}
Unsupervised domain adaptation (UDA) of time series aims to teach models to identify consistent patterns across various temporal scenarios, disregarding domain-specific differences, which can maintain their predictive accuracy and effectively adapt to new domains. However, existing UDA methods struggle to adequately extract and align both global and local features in time series data. To address this issue, we propose the \textbf{Lo}cal-\textbf{G}l\textbf{o}bal \textbf{R}epresentation \textbf{A}lignment framework (\abbr), which employs a two-branch encoder—comprising a multi-scale convolutional branch and a patching transformer branch. The encoder enables the extraction of both local and global representations from time series. A fusion module is then introduced to integrate these representations, enhancing domain-invariant feature alignment from multi-scale perspectives. To achieve effective alignment, \abbr employs strategies like invariant feature learning on the source domain, utilizing triplet loss for fine alignment and dynamic time warping-based feature alignment. Additionally, it reduces source-target domain gaps through adversarial training and per-class prototype alignment. Our evaluations on four time-series datasets demonstrate that \abbr outperforms strong baselines by up to $12.52\%$, showcasing its superiority in time series UDA tasks.  
\end{abstract}

\begin{IEEEkeywords}
Time Series Classification, Domain Adaptation.
\end{IEEEkeywords}

\section{Introduction}

Time series data is found ubiquitously across various domains, including finance, healthcare, cloud computing, and environmental monitoring \cite{koh2021wilds,wen2022robust,tang2019tensor, borges2022iot,wu2021early}. Recently, deep learning techniques have demonstrated impressive capabilities in handling various time-series datasets separately~\cite{ravuri2021skilful, jin2022multivariate, li2020efficient}. However, the challenge arises when attempting to deploy models trained on a specific source domain to tackle uncharted target domains, leading to a noticeable performance drop due to domain shifts~\cite{purushotham2016variational}. This issue underscores the critical importance of applying unsupervised domain adaptation (UDA) within the realm of time series. In the context of time series analysis, UDA aims to teach models to identify consistent patterns across various temporal scenarios, disregarding domain-specific differences, which ensures that the model can maintain its predictive accuracy and effectively adapt to new domains ~\cite{ozyurt2023contrastive}.

\begin{figure}[t]
  \centering
  
  \includegraphics[width=0.77\linewidth]{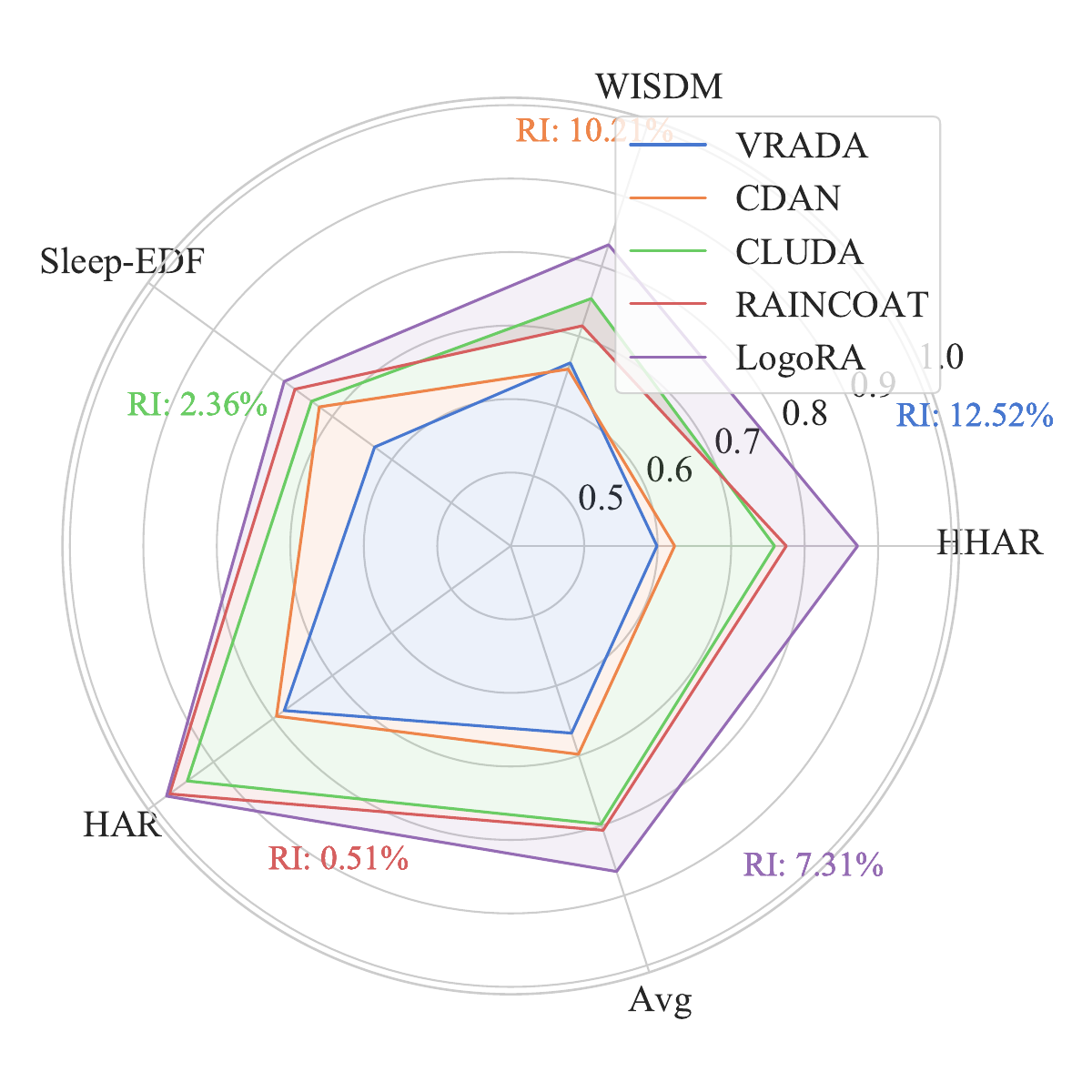}
  \vspace{-0.2cm}
  \caption{Model performance comparison on various tasks. RI denotes the relative improvement compared to SOTA.}
    \label{fig:teaser}
  \vspace{-0.3cm}  
\end{figure}

Unlike supervised approaches that rely on labeled target data, UDA leverages the wealth of information contained within the source domain and exploits it to align the distributions of source and target data in the temporal domain. Prior research endeavors in this field have employed specialized feature extractors to capture the temporal dynamics inherent in multivariate time series data. These extractors commonly rely on recurrent neural networks (RNNs)~\cite{purushotham2016variational}, long short-term memory (LSTM) networks~\cite{cai2021time}, as well as convolutional neural networks (CNNs)~\cite{liu2021adversarial,wilson2020multi,he2023domain}. Other methods~\cite{yue2022ts2vec,ozyurt2023contrastive} utilize contrastive learning to extract domain-invariant information from source domain data.

\begin{figure}[t]
  \centering
  \includegraphics[width=0.9\linewidth]{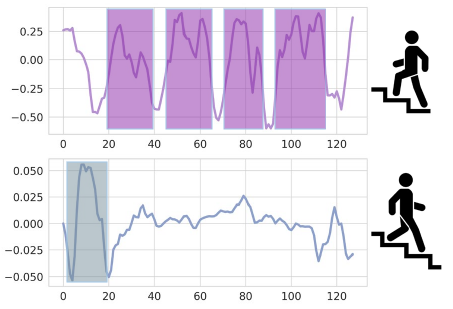}
  \vspace{-0.2cm}
  \caption{\textbf{ A motivation example}, which contains accelerometer data pieces of walking upstairs (upper) and walking downstairs (lower) from HAR dataset.}
    \label{fig:intro}
    
\end{figure}

As far as we know, none of these methods is able to adequately extract global and local features from time series data and align them across different domains. Most existing approaches~\cite{ozyurt2023contrastive} employ a temporal convolutional network (TCN)~\cite{bai2018empirical} as the backbone and use the feature of the last time step for classification. However, some intermediate time steps may contain more valuable information. As shown in the lower figure in \figurename~\ref{fig:intro}, compared to other regions, the abrupt acceleration changes in the shaded area are more representative of the class characteristics (walking downstairs) for the entire sequence, which underscores the significance of local features\footnote{Because the signal in the figure is relatively stable in other positions. If we only use the features obtained by average pooling or the features from the last time step as the classification features (like TCN), it may be unable to capture this sudden signal change. This oversight could lead to a failure in recognizing this action.}.  At the same time, focusing only on small-grained features may cause some failure cases because of ignoring long-distance feature dependencies. For instance, if only focusing on the shaded area of the upper figure in \figurename~\ref{fig:intro}, it is hard to distinguish the two classes (walking downstairs and walking upstairs), as both sequences have similar local features. While it is much easier to differentiate them through modeling the temporal dependencies between local features. Therefore, it is crucial to capture both global contextual features as well as local features in order to extract the discriminative features from time series.

In this paper, we propose a novel framework for unsupervised domain adaptation of time series data, called \abbr. Considering the network architecture, we employ a two-branch encoder, using a multi-scale convolutional branch and a patching transformer branch. The convolutional network can learn local features through convolutions~\cite{cui2016multi} and the transformer reflects long-distance feature dependencies by self-attention mechanism~\cite{zhou2021informer}. Therefore, the \abbr is able to extract local and global representations from time series instances. We then introduce a fusion module to integrate local and global representations and make the final feature for time series classification more representative and discriminative. 
 
With these representations, \abbr can better align features across different domains from multi-scale perspectives. We design the following strategies to achieve this goal: (1) \textit{invariant feature learning on source domain}: we first align patch embeddings by introducing a shortest path loss based on Dynamic Time Warping (DTW)~\cite{muller2007dynamic}, where the alignment strategy can enable the final feature more robust to time-step shift; then we employ triplet loss for finer alignment of the fused classification features of each class;  (2) \textit{reducing source-target domain gaps}: at domain level, we minimize the domain discrepancy between source and target domains through adversarial training; at class level, we introduce a per-class center loss, which reduces the distance between target domain samples and its nearest source domain prototype. \textbf{The contributions are:}

1. We propose a novel network architecture (\abbr) comprising a multi-scale local encoder utilizing a convolutional network with different kernel sizes, a global encoder employing a transformer, and a fusion module. As far as we know, the \abbr is the first UDA structure that learns a contextual representation considering both local and global patterns.

2. We design a new metric learning method based on DTW, which can overcome the severe time-shift patterns that exist in time series data and learn more robust features from the source domain. Besides, we employ adversarial learning and introduce per-class prototype-alignment strategies to align representations between the target domain and source domain, in order to acquire domain-invariant contextual information.

3. We assess the performance of \abbr on four time-series datasets: HHAR, WISDM, HAR, and Sleep-EDF. Our approach consistently outperforms state-of-the-art (SOTA) baselines by an average of $7.51\%$, as demonstrated in Figure~\ref{fig:teaser}. Furthermore, extensive empirical results validate the efficacy of each design choice in our time series UDA task. We provide insightful visual analyses to elucidate the factors contributing to the success of our algorithm.

\section{Related Work}
\textbf{Unsupervised Domain Adaptation:} Unsupervised Domain Adaptation (UDA) has witnessed substantial progress in recent years, which leverages labeled source domain to predict the labels of an unlabeled target domain. UDA methods attempt to minimize the domain discrepancy in order to lower the bound of the target error~\cite{ben2010theory}. We organize these methods into three categories: (1) \textit{Adversarial-based methods:} Adversarial training approaches aim to reduce domain shift by introducing a domain discriminator that encourages the model to learn domain-invariant features. Examples are DANN~\cite{ganin2016domain}, CDAN ~\cite{long2018conditional}, ADDA~\cite{tzeng2017adversarial}, DM-ADA~\cite{xu2020adversarial}, MADA~\cite{pei2018multi} and ELS~\cite{zhang2023free}. (2) \textit{Statistical divergence-based methods:} Statistical divergence-based methods focus on minimizing the distributional gap between the source and target domains. Maximum Mean Discrepancy (MMD)~\cite{rozantsev2018beyond} is a widely used metric for this purpose. Other examples are CORAL~\cite{sun2016deep}, DSAN~\cite{zhu2020deep}, HoMM~\cite{chen2020homm}, and MMDA~\cite{rahman2020minimum}. (3) \textit{Self-supvervised-based methods:} Self-supervised learning has emerged as a promising avenue for UDA, bypassing the need for labeled target data. Examples are CAN~\cite{kang2019contrastive}, CLDA~\cite{singh2021clda}, HCL~\cite{huang2021model}), and GRCL~\cite{tang2021gradient}. While these computer vision-based UDA methods can be applied to time series data, they may not sufficiently extract features from time series data. In contrast, \abbr is specifically designed for time series, simultaneously extracting both global and local features.

\begin{figure*}[t]
\begin{center}
\includegraphics[width=1.0\linewidth]{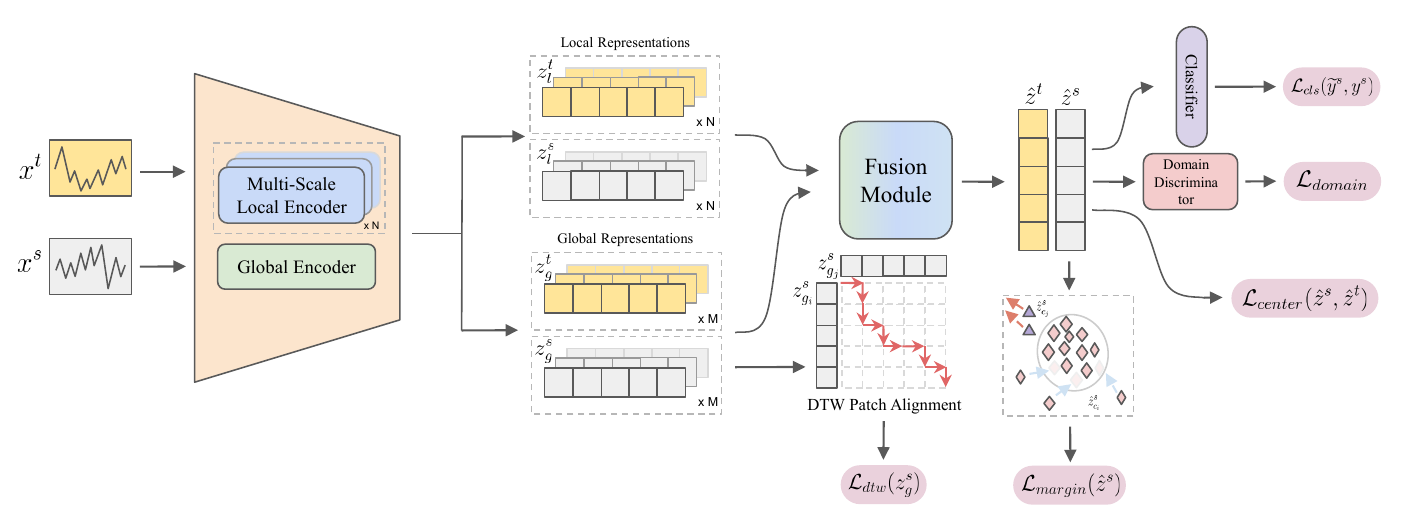}
\end{center}
\vspace{-0.3cm}
\caption{\textbf{Model Architecture and Training Pipeline of \abbr.} The time series data is processed through a feature extractor, comprising a Global Encoder and a Multi-Scale Local Encoder, to extract local and global representations. Next, these representations are fed into the Fusion Module to obtain the fused representations. We further use different representations for invariant feature learning ($\mathcal{L}_{dtw}$ and $\mathcal{L}_{margin}$) and alignment across the source and target domain ($\mathcal{L}_{domain}$ and $\mathcal{L}_{center}$).}
\label{fig:model}
\vspace{-0.2cm}
\end{figure*}

\textbf{Unsupervised Domain Adaptation for Time Series:} Despite the significant achievements of Domain Adaptation in other fields, there are only a few methods specifically tailored for time series data. (1) \textit{Adversarial-based methods:} CoDATS~\cite{wilson2020multi} builds upon the same adversarial training as VRADA, but uses a CNN for the feature extractor. (2) \textit{Statistical divergence-based methods:} SASA~\cite{cai2021time} accomplishes the alignment between the associative structure of time series variables from different domains by minimizing the maximum mean discrepancy (MMD). AdvSKM~\cite{liu2021adversarial} introduces a spectral kernel mapping to minimize MMD between different domains. RAINCOAT~\cite{he2023domain} aligns both temporal and frequency features extracted from its proposed time-frequency encoder by minimizing a domain alignment loss based on Sinkhorn divergence~\cite{cuturi2016smoothed}. (3) \textit{Self-supvervised-based methods:} DAF~\cite{jin2022domain} uses a shared attention module to extract domain-invariant and domain-specific features and then perform forecasts for source and target domains. CLUDA~\cite{ozyurt2023contrastive} applies augmentations to extract domain-invariant features by contrastive learning. However, these approaches have not thoroughly delved into the local and global characteristics present in time series data. Conversely, \abbr has the ability to deeply extract and precisely align both local and global features.

\textbf{Concept Drift and Online Time Series Forecasting:} In time series, distribution shift encompasses the phenomenon of concept drift~\cite{tsymbal2004problem}, where the underlying patterns of time series data gradually change over time. Research in this area primarily focuses on the forecasting problem~\cite{wen2024onenet}, employing techniques such as online learning~\cite{anderson2004towards} and test-time adaptation~\cite{zhang2023domain,zhang2023adanpc}. However, these methods differ from our primary interest, which centers on domain distribution shifts.

\textbf{Local and Global Feature Modeling for Time Series:} Since time series data contains rich trends and cyclical characteristics, various methods have been proposed to extract local and global features. LOGO~\cite{LOGO} compute local and global correlations within patches to represent the correlations between sensors. DeepGLO~\cite{sen2019think} is a hybrid model that combines a global matrix factorization model with another temporal network which can capture local properties of each time-series and associated covariates. To study a novel problem of multi-variates time series forecasting with missing values, LGnet~\cite{Tang_joint_2020} proposes a new framework which leverages memory network to explore global patterns given estimations from local perspectives. MICN~\cite{wang2023micn} uses down-sampled convolution and isometric convolution to extract local features and global correlations, respectively. Different from existing forecasting methods, \abbr combines transformer and CNN to capture global and multi-scale local representations of each sequence and uses cross-attention to achieve the final fine-grained feature, which presents superior performance on UDA task.

\section{Methodology}
\label{sec:method}

\subsection{Problem Definition and Overall Architecture}
\label{sec:def}
Unsupervised Domain Adaptation (UDA) addresses the challenge of transferring knowledge learned from a source domain $\mathcal{D}_S$ with labeled data to a target domain $\mathcal{D}_T$ where the label information is unavailable. We denote the source domain dataset as $\mathcal{S}=\{ (x^s_i,y^s_i) \}^{N_s}_{i=1} \sim \mathcal{D}_S$, comprising $N_s$ labeled $i.i.d.$ samples, where $x^s_i$ represents a source domain sample and $y^s_i$ is the associated label. Meanwhile, the target domain dataset is unlabeled and denoted as $\mathcal{T}=\{ (x^t_i) \}^{N_t}_{i=1} \sim \mathcal{D}_T$. 
In this paper, we focus on the classification task for multivariate time series. Hence, each sample $x_i \in \mathbb{R}^{T \times d}$ (from either the source or target domain) contains $d$ observation over $T$ time steps. During the training phase, labels for target samples $\mathcal{T}$ are inaccessible. Our aim is to learn domain-invariant and contextual information from $\mathcal{S}$ and $\mathcal{T}$.


\abbr represents a robust unsupervised approach for mining and aligning both local and global information within time series data. The complete \abbr framework for unsupervised domain adaptation is illustrated in \figurename~\ref{fig:model}. It primarily comprises four modules: a feature extractor denoted as $F(\cdot)$, a fusion module denoted as $G(\cdot)$, a classifier represented by $C(\cdot)$, and a domain discriminator labeled as $D(\cdot)$. Sec.~\ref{sec:encoder} elaborates on the specifics of the feature extractor, which utilizes source and target samples to yield global representations $z_g$ and local representations $z_l$ through the Global Encoder $F_g(\cdot)$ and the Multi-Scale Local encoder $F_l(\cdot)$. Then these representations are fed into the Local-Global Fusion Module to obtain the fused representations $\hat{z}$, as explained in Sec.~\ref{sec:fusion}. In the end, Sec.~\ref{sec:source} details the invariant feature learning on the source domain and Sec.~\ref{sec:domain_align} details the alignment across domain representations. 



\subsection{Feature Extractor}
\label{sec:encoder}

\begin{figure*}[t]
\centering  
\subfigure[\textbf{Global Encoder}]{  
\begin{minipage}[t]{0.46\linewidth}
\centering    
\includegraphics[width=0.70\linewidth]{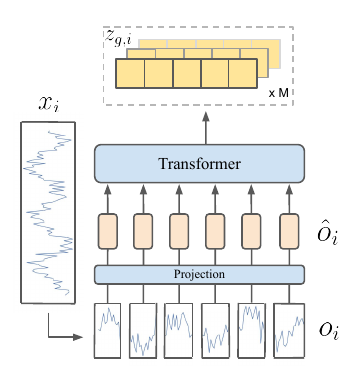} 
    \label{fig:global}
\end{minipage}
}
\subfigure[\textbf{Multi-Scale Local Encoder}]{ 
\begin{minipage}[t]{0.46\linewidth}
\centering   
\includegraphics[width=0.92\linewidth]{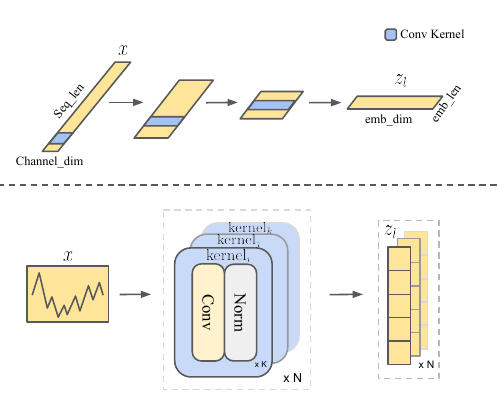}
    \label{fig:local}
\end{minipage}
}
\centering

\caption{\textbf{(a) Global Encoder:} We use a Transformer encoder with a patching operation to obtain fine-grained global representations. \textbf{(b) Multi-Scale Local Encoder:} We use ConvNet with different kernel sizes to acquire multi-scale local representations.}
\label{fig:encoder}
\end{figure*}

\begin{figure*}[h]
\begin{center}
\includegraphics[width=0.92\linewidth]{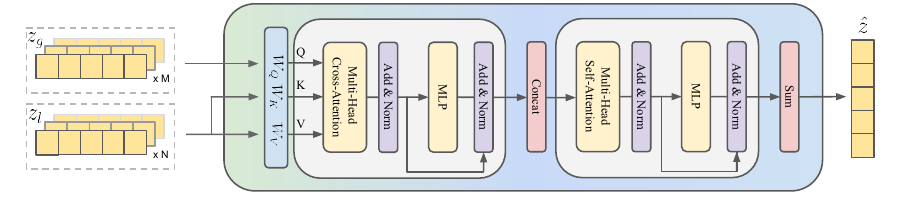}
\end{center}
\caption{\textbf{Local-Global Fusion Module.} We use cross-attention to fuse global and multi-scale local representations. Next, we concatenate all the cross-attentions and sum them to get the final output. }
\label{fig:fusion}
\end{figure*}

\textbf{Global Encoder:} 
The architecture of Global Encoder is illustrated in \figurename~\ref{fig:global}. Inspired by PatchTST~\cite{Yuqietal-2023-PatchTST}, each input multivariate time series $x_i\in \mathbb{R}^{T\times d}$ is first divided into several patches which can be either overlapped or non-overlapped. Let the patch length be denoted as $P$, and the non-overlapping stride between adjacent patches be denoted as $S$. Following the patching process, we obtain a sequence of patches $o_i \in \mathbb{R}^{M\times P\times d}$, where $M$ represents the number of patches for the original sequence $x_i$, and $ M=\lfloor\frac{(T-P)}{S} \rfloor +1$. Before patching, $S$ repeated numbers of the last time step value are padded to the end of the original sequence.

We use a vanilla Transformer encoder that maps the original sequence to the latent representations. The patches are mapped to the Transformer latent space of dimension $D$ via a trainable linear projection, resulting in $o'_{i} \in \mathbb{R}^{ M\times P\times D}$. Next, self-attention is computed within each patch and then aggregated, yielding $o''_{i} \in \mathbb{R}^ {M\times D}$. Besides, a learnable position encoding $W_{pos} \in \mathbb{R}^{M\times D}$ is also applied to monitor the temporal order of patches to get patch embeddings: $\hat{o}_{i}=o''_{i} +W_{pos}$, each of which represents a segment of the original sequence. Finally, the patch embeddings $\hat{o}_{i}$ are fed into Transformer to generate the global representations $z_{g,i} \in \mathbb{R}^{M\times D}$.



\textbf{Multi-Scale Local Encoder:} 
The Multi-Scale Local Encoder comprises $N$ convolutional neural networks (CNNs) with varying kernel sizes ($Kernel_i, i=1, ..., N$), where each CNN consists of $K$ stages. As depicted in the lower part of \figurename~\ref{fig:local}, each stage consists of a convolution operation followed by batch normalization. With each stage, a shorter yet deeper embedding is obtained to facilitate the preservation of more comprehensive local information, which is shown in the upper part of \figurename~\ref{fig:local}. In summary, given an input time series $x\in \mathbb{R}^{T\times d}$, the Multi-Scale Local Encoder generates $N$ local representations $z_l = \{z_l^{(i)}\in \mathbb{R}^{l^{(i)}_\text{emb}\times d_\text{emb}} \ | \ i=1, ..., N\}$, where $l_\text{emb}$ and $d_\text{emb}$ represent the length and dimension of the output embedding from the final stage.

\subsection{Local-Global Fusion Module}
\label{sec:fusion}

Given the local and global representations $z_l, z_g$ of the time series instances, we propose the Local-Global Fusion Module to further integrate them into unified representations, as depicted in \figurename~\ref{fig:fusion}. 

The core of the fusion module is the cross-attention mechanism. Similar to self-attention, the input $z_g$ and $z_l$ are first projected into three vectors respectively, $i.e.$ queries $Q\in \mathbb{R}^{M\times d_k}$, keys $K \in \mathbb{R}^{l_\text{emb}\times d_k}$ and values $V\in \mathbb{R}^{l_\text{emb} \times d_v}$, where $d_k$ and $d_v$ indicate the dimensions of them. Notably, for each local representation $z_l^{(i)}$ in $z_l$, there exits a corresponding key $K^{(i)}\in \mathbb{R}^{l_\text{emb}^{(i)}\times d_k}$ and value $V^{(i)}\in \mathbb{R}^{l_\text{emb}^{(i)} \times d_v}$. In order to maximize the incorporation of global information and diverse scale local information, we compute the cross-attention between global representations $z_g$ and each local representation $z_l^{(i)}$. The specific calculation process is as follows:
\begin{equation}
    Attn_{cross}^{(i)}(Q,K^{(i)},V^{(i)})=\text{Softmax}(\frac{Q{K^{(i)}}^T}{\sqrt{d_k}})V^{(i)}.
\end{equation}
The output $Attn_{cross} = \{ Attn_{cross}^{(i)} \in \mathbb{R}^{M\times d_v} \ | \ i = 1,...N\}$ holds the same length $M$ as the number of the queries. Next, we concatenate all the cross-attention $Attn_{cross}^{(i)}$ corresponding to different scales and get a feature of length $N\cdot M$ and dimensionality $d_v$. Finally, the fused contextual representation $\hat{z}$, which combines both global and local information, is obtained by computing self-attention on the concatenated feature and subsequently summing the results. 

\subsection{Invariant Feature Learning On Source Domain}
\label{sec:source}


\begin{figure}[t]
  \centering
  \includegraphics[width=1.0\linewidth]{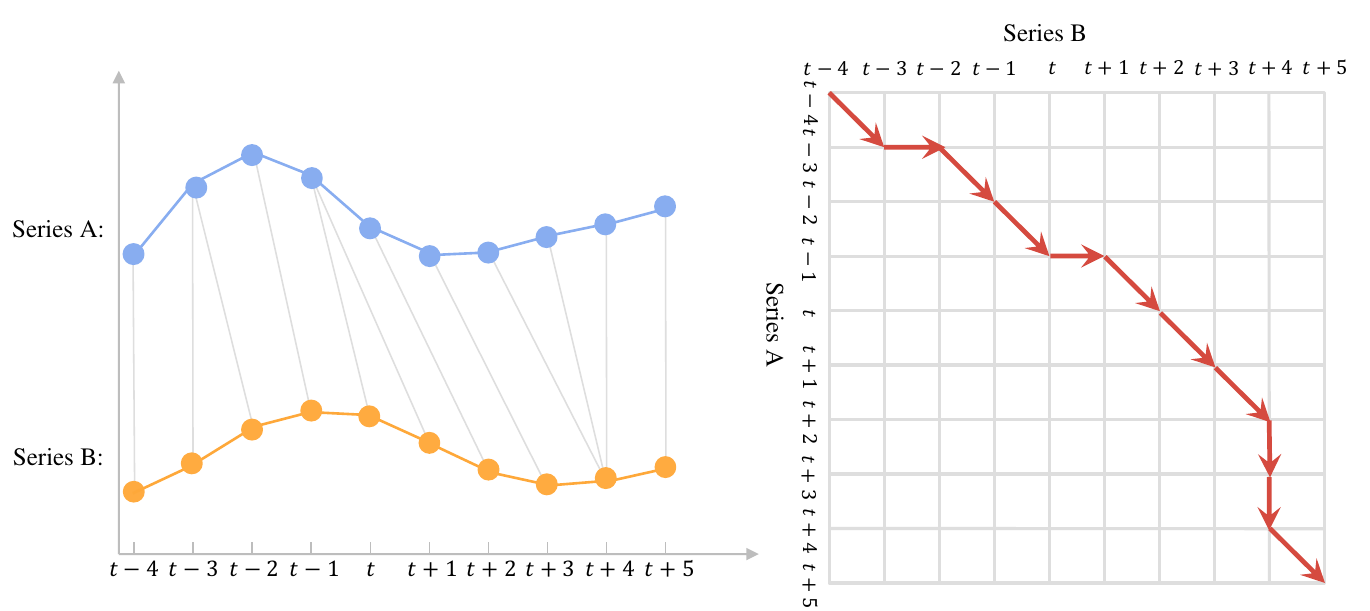}
  \caption{\textbf{Example of local distance computed by DTW.} The grey lines on the upper show the corresponding alignment between two segments of time series and the red arrows on the right show the shortest path in the distance matrix.}
    \label{fig:dtw}
\end{figure}

Time shifts are particularly common between different sequences in time series data~\cite{cai2021time}, as depicted in \figurename~\ref{fig:dtw}. In such cases, traditional Euclidean distance may not accurately measure the similarity between two sequences. Instead, the Dynamic Time Warping (DTW) algorithm~\cite{muller2007dynamic} is well-suited for calculating distances between time series with time-step shifts. However, directly calculating DTW leads to a significant increase in time complexity. Fortunately, with the patching operation in Sec.~\ref{sec:encoder}, we can perform calculations at the patch level. Thus, to ensure the learned representation is robust to time-step shift, we align the patch representations based on the DTW distance matrix. Specifically, we propose the \textit{DTW Alignment Loss} $\mathcal{L}_{dtw}$ on $z_g^s$ as follows:
\begin{equation}
    \mathcal{L}_{dtw}=\sum\limits_{i}^{N_s} \text{max}(\text{DTW}(z_{g,i}^s,p)-\text{DTW}(z_{g,i}^s,n)+\alpha,0),
\end{equation}
where $z_{g,i}^s$ is an anchor input, $p$ is a randomly selected positive input of the same class as $z_{g,i}^s$, $n$ is a randomly selected negative input of a different class from $z_{g,i}^s$, $\alpha$ is a margin between positive and negative pairs, and $\text{DTW}(\cdot)$ represents the DTW distance. After alignment with $\mathcal{L}_{dtw}$, the output obtained from the fusion of global and local representations is also robust to time-shift. To further enhance the discrimination capability and fully leverage the label information on the source domain, we align the final fused representation by the proposed \textit{Margin Alignment Loss} $\mathcal{L}_{margin}$ based on triplet loss. The specific definition is as follows:
\begin{equation}
    \mathcal{L}_{margin}=\sum\limits_{i}^{N_s} \text{max}(\text{dist}(\hat{z}_i^s,p)-\text{dist}(\hat{z}_i^s,n)+\beta,0),
\end{equation}
where $\hat{z}_i^s$ is an anchor input, $p$ is a randomly selected positive input of the same class as $\hat{z}_i^s$, $n$ is a randomly selected negative input of a different class from $\hat{z}_i^s$, $\beta$ is a margin between positive and negative pairs, and $\text{dist}(\cdot)$ represents the Euclidean distance.

\subsection{Alignment Across Domain Representations}
\label{sec:domain_align}

Borrowing the idea from DANN~\cite{ganin2016domain}, we employ adversarial training for unsupervised domain adaptation by minimizing a combination of two losses. The first one is the \textit{classification loss} $\mathcal{L}_{cls}$, which trains the feature extractor $F(\cdot)$, the Local-global Fusion Module $G(\cdot)$, and the classifier $C(\cdot)$ with the data from the source domain. The specific definition is as follows:

\begin{equation}
    \mathcal{L}_{cls}=\frac{1}{N_s} \sum\limits_{i}^{N_s} \text{CrossEntropy}(C(G(F(x^s_i))),y^s_i).
\end{equation}

Since $\mathcal{L}_{cls}$ is minimized to guarantee lower source risk, another loss $\mathcal{L}_{domain}$ is minimized over the domain discriminator $D(\cdot)$ but maximized over $F(\cdot)$, $G(\cdot)$ and $C(\cdot)$:

\begin{equation}
\begin{split}
     \mathcal{L}_{domain} &=\frac{1}{N_s} \sum\limits_{i}^{N_s} \log[D(G(F(x^s_i)))]\\
            & -\frac{1}{N_t} \sum\limits_{i}^{N_t} \log[1-D(G(F(x^t_i)))].
\end{split}
\end{equation}

Then the complete minmax game of adversarial training in \abbr is:
\begin{equation}
    \min\limits_{C,F,G} \mathcal{L}_{cls}-\lambda_{domain}\mathcal{L}_{domain};\quad
    \min\limits_{D} \mathcal{L}_{domain}.
\end{equation}

This dual optimization aims to model source and target domain features within the same feature space, with the ultimate objective of learning domain-invariant knowledge. And by enforcing the order of distances, $\mathcal{L}_{margin}$ and $\mathcal{L}_{dtw}$ model embeddings with the same labels closer than those with different labels in the feature space. Consequently, we further propose the \textit{center loss} $\mathcal{L}_{center}$ to align features of the same class between the target domain and the source domain. This is accomplished by reducing the distance between target domain samples and corresponding source domain prototypes. The definition is as follows:
\begin{equation}
    \mathcal{L}_{center}=\sum\limits_{i}^{N_t} \underset{j}{\text{min}} ||\hat{z}_i^t-c_j||^2 ,
\end{equation}
where $c_j$ is the $j$th class prototype in the source domain. 

\subsection{Training}
In summary, the overall loss of \abbr framework is
\begin{equation}
\begin{split}
     \mathcal{L}_{total} = &\mathcal{L}_{cls}- \lambda_{domain}\mathcal{L}_{domain}+\lambda_{margin}\mathcal{L}_{margin} \\
    &+\lambda_{dtw}\mathcal{L}_{dtw}+\lambda_{center}\mathcal{L}_{center} ,
\end{split}
\end{equation}
where hyper-parameters $\lambda_{domain},\lambda_{margin},\lambda_{dtw}$ and $\lambda_{center}$ control the contribution of each component.
And the complete training follows:
\begin{equation}
\begin{aligned}
    \min\limits_{C,F,G}\mathcal{L}_{total}; 
    \min\limits_D &\mathcal{L}_{domain} .
\end{aligned}   
\end{equation}
The detailed training algorithm is depicted in Alg.~\ref{alg:main}.

\begin{algorithm}[t]
   \caption{Training and inference algorithm of \abbr}
   \label{alg:main}
\begin{algorithmic}[1]
   \STATE {\bfseries Input:} Multivariate time series $x \in \mathbb{R}^{T \times d}$ with $d$ variables and length $T$, where $x^s$ represents the data from source domain and $x^t$ represents the data from target domain. Only the label of source domain $y^t$ is accessible. The entire framework is trained for $E$ epochs.
   \STATE {\bfseries Initialize:}  a global encoder $f_{g}$, a multi-scale local encoder $f_{l}$, a fusion module $g$, a classifier $f_{cls}$ and a domain discriminator $f_{domain}$.
   \FOR{each $i \in [1,E]$}
       \STATE {\color{brown}\bfseries Get representations from two encoders.}
       \STATE $z_g\in \mathbb{R}^{M\times D}=f_g(x)$  \\
       \textit{ // Global representations from global encoder}\\
       \STATE $z_l = \{z_l^{(i)}\in \mathbb{R}^{l^{(i)}_\text{emb}\times d_\text{emb}} \ | \ i=1, ..., N\}=f_l(x)$  \\
       \textit{ $\;$// Local representations from local encoder}\\
       \STATE {\color{brown}\bfseries Get fused representations from fusion module.}
       \STATE $\hat{z}\in \mathbb{R}^{D}=g(x)$  \\
       \textit{ // Fused representations from fusion module}\\
       \STATE {\color{brown}\bfseries Get prediction results from classifier.}
       \STATE $\widetilde{y^s}\in \mathbb{R}^{C}=f_{cls}(\hat{z}^s)$  \\
       \textit{ // Prediction results on source domain}\\
       \STATE $\widetilde{y^t}\in \mathbb{R}^{C}=f_{cls}(\hat{z}^t)$  \\
       \textit{ // Prediction results on target domain}\\
       \STATE {\color{brown}\bfseries Adversarial training.}
       \STATE Compute: $\mathcal{L}_{domain}$
       \STATE Update $f_{domain}$ with $\nabla\mathcal{L}_{domain}$
       
       \STATE {\color{brown}\bfseries Invariant feature learning and alignment across domains}
       \STATE Compute: $\mathcal{L}_{cls}$, $\mathcal{L}_{domain}$, $\mathcal{L}_{global}$, $\mathcal{L}_{dtw}$, $\mathcal{L}_{center}$
       \STATE $\mathcal{L}_{total}=\mathcal{L}_{cls}- \lambda_{domain}\mathcal{L}_{domain}+\lambda_{global}\mathcal{L}_{global}$\\
       $\quad\quad\quad\quad\quad +\lambda_{dtw}\mathcal{L}_{dtw}+\lambda_{center}\mathcal{L}_{center}$
       \STATE Update $f_{g}$, $f_{l}$, $g$, $f_{cls}$ with $\nabla\mathcal{L}_{total}$
       
\ENDFOR     
\end{algorithmic}
\end{algorithm}

\section{Experiments}
\label{sec:exp}
\subsection{Experimental Setup}

\textbf{Dataset:} We consider four benchmark datasets from  multiple modalities: WISDM~\cite{kwapisz2011activity}, HAR~\cite{anguita2013public}, HHAR~\cite{stisen2015smart}, Sleep-EDF~\cite{goldberger2000physiobank}. The details are as follows:

(1) HHAR~\cite{stisen2015smart}: The dataset comprises 3-axis accelerometer measurements from 30 participants. These measurements were recorded at 50 Hz, and we utilize non-overlapping segments of 128 time steps to predict the participant's activity type. The activities fall into six categories: biking, sitting, standing, walking, walking upstairs, and walking downstairs.

(2) WISDM~\cite{kwapisz2011activity}: The dataset comprises 3-axis accelerometer measurements collected from 30 participants at a frequency of 20 Hz. To predict the activity label of each participant during specific time segments, we employ non-overlapping segments consisting of 128 time steps. The dataset encompasses six distinct activity labels: walking, jogging, sitting, standing, walking upstairs, and walking downstairs.

(3) HAR~\cite{anguita2013public}: The dataset encompasses measurements from a 3-axis accelerometer, 3-axis gyroscope, and 3-axis body acceleration. This data is gathered from 30 participants at a sampling rate of 50 Hz. Like the WISDM dataset, we employ non-overlapping segments of 128 time steps for classification. The objective is to classify the time series into six activities: walking, walking upstairs, walking downstairs, sitting, standing, and lying down.

(4) Sleep-EDF~\cite{goldberger2000physiobank}: The dataset comprises electroencephalography (EEG) readings from 20 healthy individuals. The goal is to classify the EEG readings into five sleep stages: wake (W), non-rapid eye movement stages (N1, N2, N3), and rapid eye movement (REM). Consistent with previous research, our analysis primarily focuses on the Fpz-Cz channel.

\begin{table}[t]
\caption{\textbf{Details of benchmark datasets.}}  
\centering
\resizebox{0.98\columnwidth}{!}{%
\begin{tabular}{@{}ccccccc@{}}
\toprule
Dataset & Subjects & Channels & Length & Class & Train & Test\\
\midrule
HHAR & 9 &3 & 128 & 6 & 12,716 & 5,218\\
WISDM & 30 &3 & 128 & 6 & 1,350 & 720\\
HHAR & 30 &9 & 128 & 6 & 2,300 & 990\\
Sleep-EDF & 20 &1 & 3000 & 2 & 160,719 & 107,400\\
\bottomrule
\end{tabular}%
}
\end{table}
Following previous work on DA for time series~\cite{ozyurt2023contrastive,he2023domain}, we select the same ten pairs of domains to specify source $\mapsto$ target domains.

\textbf{Baseline:} We compare the performance of our \abbr on unsupervised domain adaptation with five state-of-the-art baselines for UDA of time series: VRADA~\cite{purushotham2016variational}, CoDATS~\cite{wilson2020multi}, AdvSKM~\cite{liu2021adversarial}, CLUDA~\cite{ozyurt2023contrastive} and RAINCOAT~\cite{he2023domain}. We also consider five general UDA methods for a comprehensive comparison: CDAN~\cite{long2018conditional}, DeepCORAL~\cite{sun2016deep}, DSAN~\cite{zhu2020deep}, HoMM~\cite{chen2020homm} and MMDA~\cite{rahman2020minimum}.

\textbf{Implementation Details:} The implementation was done in PyTorch. Except for RAINCOAT~\cite{he2023domain}, all other baselines are configured according to the experimental settings within CLUDA~\cite{ozyurt2023contrastive}. During model training, we employed the Adam optimizer for all methods, with carefully tuned learning rates specific to each method. The hyperparameters of Adam were selected after conducting a grid search on source validation datasets, exploring a range of learning rates from $1 \times 10^{-4}$ to $1 \times 10^{-2}$. The learning rates were chosen to optimize the performance of each method. For \abbr, we use an 8-layer transformer as the global encoder and three convolutional networks with varying kernel sizes as the local encoder. In the patching operation, for patches of different lengths, we uniformly use half of their length as the stride. And the stride in local encoder is always set to be 1. 

\textbf{Evaluation:}
We report the mean accuracy calculated on target test datasets. It is computed by dividing the number of correctly classified samples by the total number of samples.

\begin{table*}[t]
\centering
\caption{\textbf{UDA performance on benchmark datasets.} Prediction accuracy for each dataset between various subjects. \abbr consistently outperforms all other methods in accuracy on test sets drawn from the target domain dataset.}\label{tab:acc}
\resizebox{0.9\linewidth}{!}{%

    \begin{tabular}{cccccccccccgg}
    \toprule
    {Source $\mapsto$ Target} & {VRADA} & {CoDATS} & {AdvSKM} & {CDAN}  & {CORAL} & {DSAN} & {HoMM} & {MMDA} & {CLUDA} & {RAINCOAT} & \textbf{\abbr} & Improvement\\
     \midrule
    HHAR 0 $\mapsto$ 2 & 0.593 & 0.65 & 0.681 &  0.676 & 0.618 & 0.292 & 0.680 & 0.671 & 0.726 & 0.788 & \textbf{0.840} & +6.60$\%$\\
    HHAR 1 $\mapsto$ 6 & 0.690 & 0.686 & 0.652 & 0.717 & 0.712 & 0.689 & 0.725 & 0.686 & 0.855 & 0.889 & \textbf{0.916} & +3.04$\%$\\
    HHAR 2 $\mapsto$ 4 & 0.476 & 0.381 & 0.291 & 0.472 & 0.332 & 0.229 & 0.332 & 0.238 & 0.585 & 0.538 & \textbf{0.936} & +60.00$\%$\\
    HHAR 4 $\mapsto$ 0 & 0.263 & 0.229 & 0.203 & 0.262 & 0.259 & 0.193 & 0.193 & 0.205 & 0.353 & 0.268 & \textbf{0.389} & +10.20$\%$\\
    HHAR 4 $\mapsto$ 1 & 0.558 & 0.501 & 0.494 & 0.690 & 0.482 & 0.504 & 0.628 & 0.551 & 0.774 & 0.898 & \textbf{0.963} & +7.24$\%$\\
    HHAR 5 $\mapsto$ 1 & 0.775 & 0.761 & 0.737 & 0.857 & 0.787 & 0.407 & 0.787 & 0.790 & 0.948 & 0.977 &  \textbf{0.985} & +0.82$\%$\\
    HHAR 7 $\mapsto$ 1 & 0.575 & 0.551 & 0.426 & 0.413 & 0.511 & 0.366 & 0.496 & 0.415 & 0.875 & 0.887 & \textbf{0.948} & +6.88$\%$\\
    HHAR 7 $\mapsto$ 5 & 0.523 & 0.380 & 0.192 & 0.492 & 0.489 & 0.233 & 0.328 & 0.320 & 0.636 & \textbf{0.852} & 0.815 & -0.34$\%$\\
    HHAR 8 $\mapsto$ 3 & 0.813 & 0.766 & 0.748 & 0.942 & 0.869 & 0.602 & 0.844 & 0.934 & 0.942 & 0.973 & \textbf{0.974} & +0.10$\%$\\
    HHAR 8 $\mapsto$ 4 & 0.720 & 0.601 & 0.650 & 0.712 & 0.618 & 0.516 & 0.658 & 0.701 & 0.896 & 0.796 & \textbf{0.968} & +8.04$\%$\\ 
    \midrule
    HHAR Avg & 0.599 & 0.551 & 0.508 & 0.623 & 0.568 & 0.403 & 0.567 & 0.551 & 0.759 & 0.775 &  \textbf{0.872} & +12.52$\%$\\
    \midrule
    WISDM 12 $\mapsto$ 19 & 0.558 & 0.633 & 0.639 & 0.488 & 0.433 & 0.639 & 0.415 & 0.358 & 0.694 & 0.530 &  \textbf{0.742} & +6.92$\%$\\
    WISDM 12 $\mapsto$ 7 & 0.708 & 0.721 & 0.742 & 0.771 & 0.592 & 0.625 & 0.546 & 0.679 & 0.792 & 0.875 & \textbf{0.896} & +2.40$\%$\\
    WISDM 18 $\mapsto$ 20 & 0.571 & 0.634 & 0.390 & 0.771 & 0.380 & 0.366 & 0.429 & 0.380 & 0.780 & 0.756 & \textbf{0.829} & +6.28$\%$\\
    WISDM 19 $\mapsto$ 2 & 0.644 & 0.395 & 0.434 & 0.346 & 0.473 & 0.366 & 0.488 & 0.385 & 0.561 & 0.659 & \textbf{0.756} & +14.72$\%$\\
    WISDM 2 $\mapsto$ 28 & 0.729 & 0.809 & 0.809 & 0.813 & 0.827 & 0.773 & 0.787 & 0.813 & 0.849 & 0.798 & \textbf{0.889} & +4.71$\%$\\
    WISDM 26 $\mapsto$ 2 & 0.683 & 0.727 & 0.620 & 0.615 & 0.737 & 0.605 & 0.702 & 0.634 & 0.863 &  0.598 & \textbf{0.878} & +1.74$\%$\\
    WISDM 28 $\mapsto$ 2 & 0.688 & 0.717 & 0.707 & 0.580 & 0.649 & 0.673 & 0.644 & 0.668 & 0.741 & 0.585 & \textbf{0.854} & +15.25$\%$\\
    WISDM 28 $\mapsto$ 20 & 0.741 & 0.741  & 0.707 & 0.776 & 0.737 & 0.746 & 0.790 & 0.722 & 0.820 & 0.804 & \textbf{0.927} & +13.05$\%$\\
    WISDM 7 $\mapsto$ 2 & 0.605 & 0.610 & 0.610 & 0.649 & 0.624 & 0.620 & 0.605 & 0.605 & 0.712 & \textbf{0.817} & 0.781 & -4.41$\%$\\
    WISDM 7 $\mapsto$ 26 & 0.693 & 0.702 & 0.702 & 0.722 & 0.683 & 0.698 & 0.698 & 0.712 & 0.727 & 0.732 & \textbf{0.756} & +3.28$\%$\\ 
    \midrule
    WISDM Avg & 0.662 & 0.669 & 0.636 & 0.653 & 0.713 & 0.611 & 0.610 & 0.596 & 0.754 & 0.715 &  \textbf{0.831} & +10.21$\%$\\
    \midrule
    Sleep-EDF 0 $\mapsto$ 11 & 0.499 & 0.695 & 0.565 & 0.689 & 0.572 & 0.518 & 0.278 & 0.245 & 0.579 & 0.744 &  \textbf{0.746} & +0.27$\%$\\
    Sleep-EDF 2 $\mapsto$ 5 & 0.578 & 0.718 & 0.656 & 0.695 & 0.604 & 0.422 & 0.455 & 0.438 & 0.719 & 0.738 & \textbf{0.754} & +2.17$\%$\\
    Sleep-EDF 12 $\mapsto$ 5 & 0.655 & 0.793 & 0.765 & 0.785 & 0.750 & 0.434 & 0.399 & 0.471 & 0.794 & 0.798 & \textbf{0.849} & +6.39$\%$\\
    Sleep-EDF 7 $\mapsto$ 18 & 0.671 & 0.732 & 0.609 & 0.732 & 0.658 & 0.360 & 0.549 & 0.533 & 0.745 & 0.753 & \textbf{0.770} & +2.26$\%$\\
    Sleep-EDF 16 $\mapsto$ 1 & 0.798 & 0.753 & 0.730 & 0.745 & 0.695 & 0.534 & 0.507 & 0.547& 0.758 & \textbf{0.786} & 0.748 &  $-7.99\%$\\
    Sleep-EDF 9 $\mapsto$ 14 & 0.733 & 0.816 & 0.768 & 0.801 & 0.822 & 0.503 & 0.469 & 0.504 & 0.863 &  \textbf{0.872} & 0.862 & -1.15$\%$\\
    Sleep-EDF 4 $\mapsto$ 12 & 0.576 & \textbf{0.717} & 0.661 & 0.671 & 0.415 & 0.424 & 0.435 & 0.668 & 0.665 & 0.699 & 0.688 & -4.04$\%$\\
    Sleep-EDF 10 $\mapsto$ 7 & 0.570 & 0.733  & 0.743 & 0.734 & 0.761 & 0.529 & 0.517 & 0.526 & 0.752 & 0.772 & \textbf{0.797} & +3.24$\%$\\
    Sleep-EDF 6 $\mapsto$ 3 & 0.751 & 0.836 & 0.789 & 0.810 & 0.784 & 0.531 & 0.510 & 0.506 & 0.820 & \textbf{0.846} & 0.841 & -0.59$\%$\\
    Sleep-EDF 8 $\mapsto$ 10 & 0.458 & 0.442 & 0.448 & 0.552 & 0.368 & 0.544 & 0.501 & 0.436 & 0.657 & 0.624 & \textbf{0.754} & +5.90$\%$\\ 
    \midrule
     Sleep-EDF Avg & 0.629 & 0.724 & 0.673 & 0.722 & 0.667 & 0.461 & 0.479 & 0.464 & 0.735 & 0.763 &  \textbf{0.781} & +2.36$\%$\\
    \midrule
    HAR 15 $\mapsto$ 19 & 0.756 & 0.733 & 0.741 & 0.759 & 0.759 & 0.874 & 0.748 & 0.726 & 0.967 & 1.000 & \textbf{1.000} & +0.00$\%$\\
    HAR 18 $\mapsto$ 21 & 0.794 & 0.522 & 0.555 & 0.803 & 0.610 & 0.558 & 0.581 & 0.555 & 0.910 & 1.000 & \textbf{1.000} & +0.00$\%$\\
    HAR 19 $\mapsto$ 25 & 0.768 & 0.468 & 0.452 & 0.771 & 0.590 & 0.774 & 0.487 & 0.448 & \textbf{0.932} & 0.885 & 0.887 & -4.83$\%$\\
    HAR 19 $\mapsto$ 27 & 0.793 & 0.709 & 0.723 & 0.807 & 0.744 & 0.891 & 0.726 & 0.754 & 0.996 & 0.989 & \textbf{1.000} & +0.40$\%$\\
    HAR 20 $\mapsto$ 6 & 0.808 & 0.661 & 0.641 & 0.820 & 0.686 & 0.784 & 0.673 & 0.694 & 1.000 & 0.989 & \textbf{1.000} & +0.00$\%$\\
    HAR 23 $\mapsto$ 13 & 0.736 & 0.504 & 0.504 & 0.700 & 0.688 & 0.628 & 0.604 & 0.572 & 0.778 & 0.885 &  \textbf{0.920} & +3.62$\%$\\
    HAR 24 $\mapsto$ 22 & 0.837 & 0.820 & 0.833 & 0.837 & 0.743 & 0.808 & 0.853 & 0.829 & 0.988 & 1.000 & \textbf{1.000} & +0.00$\%$\\
    HAR 25 $\mapsto$ 24 & 0.817 & 0.583 & 0.566 & 0.790 & 0.648 & 0.883 & 0.607 & 0.666 & 0.993 & 1.000 & \textbf{1.000} & +0.00$\%$\\
    HAR 3 $\mapsto$ 20 & 0.752 & 0.874 & 0.878 & 0.815& 0.848 & 0.804 & 0.874 & 0.815 & 0.967 & \textbf{1.000} & 0.982 & -0.18$\%$\\
    HAR 13 $\mapsto$ 19 & 0.752 & 0.793 & 0.807 & 0.841 & 0.793 & 0.726 & 0.815 & 0.800 & 0.967 & 1.000 & \textbf{1.000} & +0.00$\%$\\ 
    \midrule
    HAR Avg & 0.781 & 0.670 & 0.670 & 0.794 & 0.709 & 0.773 & 0.697 & 0.686 & 0.944 & 0.974 &  \textbf{0.979} & +0.51$\%$\\
    \bottomrule
    \end{tabular}%
}
\end{table*}

\subsection{Numerical Results on UDA Benchmarks}
\label{sec:uda}

Following previous work, we present the prediction results for 10 source-target domain pairs for each dataset in Table~\ref{tab:acc}. Overall, \abbr has won 4 out of 4 datasets and achieves an average improvement of accuracy ($6.40\%$) over the strongest baseline. Specifically, on HHAR dataset, our \abbr outperforms the best baseline accuracy of RAINCOAT by $12.52\%$ (0.872 vs. 0.775). On WISDM dataset, our \abbr outperforms the best baseline accuracy of CLUDA by $10.21 \%$ (0.831 vs. 0.754). On Sleep-EDF dataset, our \abbr outperforms the best baseline accuracy of RAINCOAT by $2.36 \%$ (0.781 vs. 0.763). And on HAR dataset, although the current state-of-the-art RAINCOAT achieves 0.974 in terms of mean accuracy, our work pushes that boundary without 2-stage training or any correction steps, achieving a $0.51\%$ (0.979 vs. 0.974) absolute improvement over RAINCOAT and 3.71\% (0.979 vs. 0.944) absolute improvement over CLUDA. The overall performance demonstrates that \abbr successfully extracts both global and local information from sequences, learning domain-invariant features and achieving alignment across different domains. This enhances knowledge transfer for time series data in the presence of domain shifts.

\subsection{Ablation Studies}
\label{sec: ablation}


\begin{table*}[t]
    \centering
    \caption{\textbf{Ablation studies of loss function.} Specifically, the loss functions, $\mathcal{L}_{cls}$, $\mathcal{L}_{domain}$, $\mathcal{L}_{margin}$, $\mathcal{L}_{dtw}$, and $\mathcal{L}_{center}$, are shown below. The first column presents results using TCN along with our loss functions. When only the classification loss $\mathcal{L}_{cls}$ is used (second column), it refers to a source-only model, which is trained exclusively on the source domain. We evaluate \abbr across 10 scenarios on all the datasets and report the mean Accuracy.}
    \label{tab:ablation}
    \begin{tabular}{c|cccccc|c}
    \toprule
        Dataset & TCN w/ alignment & w/o alignment & w/o $\mathcal{L}_{domain}$ & w/o $\mathcal{L}_{center}$ & w/o $\mathcal{L}_{margin}$ & w/o $\mathcal{L}_{dtw}$ & \abbr\\
        \midrule
        HHAR & 0.721 & 0.707 & 0.734 & 0.795 & 0.764 & 0.787 & \textbf{0.872}\\
        WISDM & 0.709 & 0.712 & 0.741 & 0.799 & 0.813 & 0.794 & \textbf{0.831}\\
        Sleep-EDF & 0.758 & 0.674 & 0.703 & 0.760 & 0.757 & 0.751 & \textbf{0.781}\\
        HAR & 0.892 & 0.877 & 0.898 & 0.952 & 0.911 & 0.936 & \textbf{0.979}\\
        \bottomrule
    \end{tabular}
    
\end{table*}

\begin{figure*}[t]
\begin{center}
\includegraphics[width=0.85\linewidth]{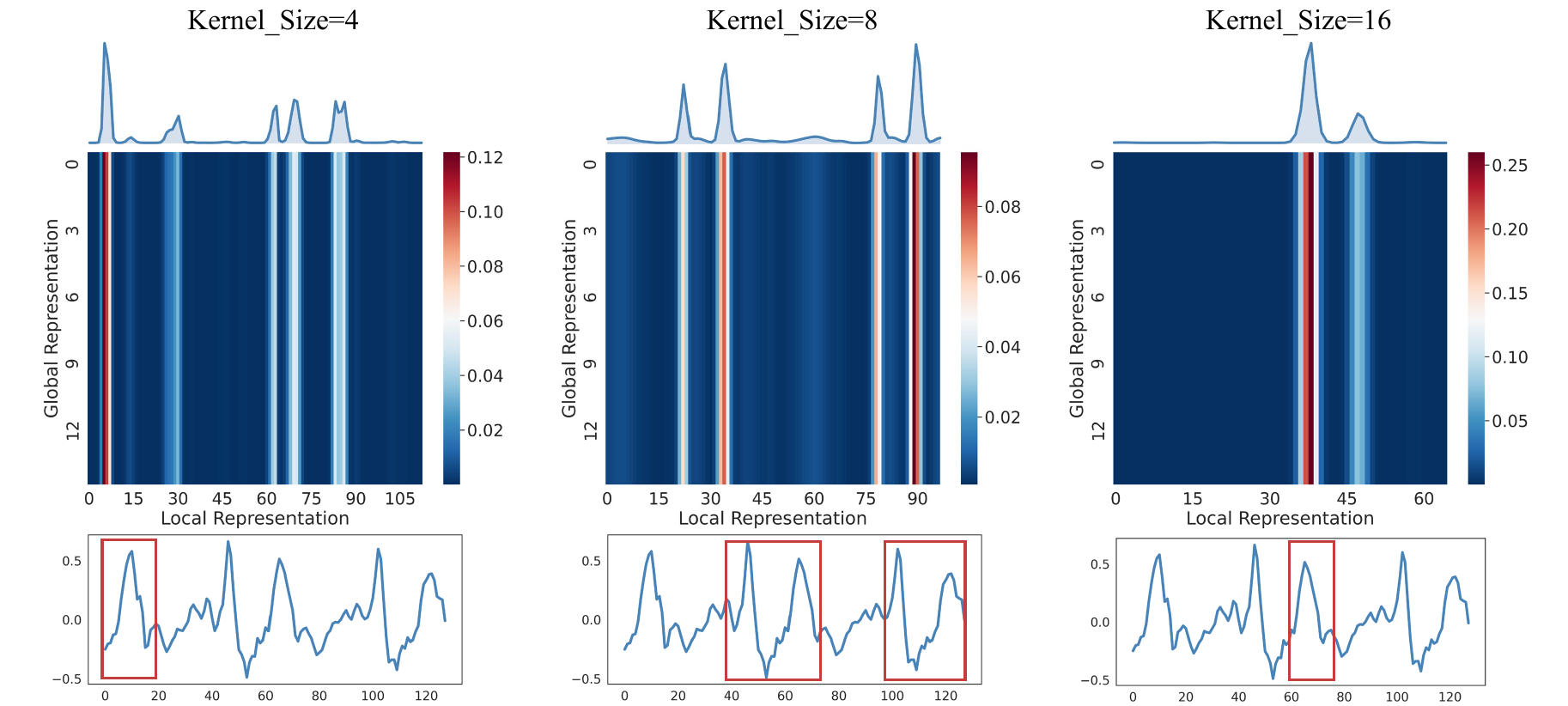}
\end{center}
\caption{\textbf{Cross-Attention Heat Map:} The heatmap displays the cross-attention weights calculated between global representation and multi-scale local representation within the fusion module. The horizontal and vertical axes represent the lengths of the local and global representation. From left to right, the results are shown for kernel sizes of 4, 8, and 16, respectively. The figure below represents the specific data from one channel of the original time series.}
\label{fig:attn}
\end{figure*}

\begin{table}[t]
\centering
\caption{\textbf{Ablation studies of model architecture.} We evaluate different backbones across 10 scenarios on the HHAR dataset and report the Accuracy. All the models are only trained to minimize the classification loss on the source domain.}\label{tab:ab_model}

\resizebox{\columnwidth}{!}{%
\begin{tabular}{@{}cccccc@{}}
\toprule
\multirow{2}{*}{Model} & \multicolumn{5}{c}{UDA performance} \\  \cmidrule(lr){2-6}
 & 2 $\mapsto$ 4 & 4 $\mapsto$ 1 & 7 $\mapsto$ 1 & 8 $\mapsto$ 3 & Avg (10 scenarios) \\
\midrule
TCN & 0.296 & 0.454 & 0.358 & 0.358 & 0.496 \\
 Transformer & 0.502 & 0.563 & 0.567 & 0.852 & 0.601 \\
 PatchTST & 0.442 & 0.683 & 0.731 & 0.834 & 0.629 \\ 
Transformer $+$ TCN & 0.463 & 0.689 & 0.610 & 0.659 & 0.619 \\ 
Global Encoder $+$ TCN & 0.346 & 0.560 & 0.739 & 0.865 & 0.623 \\ 
Global Encoder $+$ Local Encoder & 0.406 & 0.716 & 0.582 & 0.825 & 0.633 \\ 
 \abbr & \textbf{0.514} & \textbf{0.772} & \textbf{0.743} & \textbf{0.891} & \textbf{0.707} \\
\bottomrule
\end{tabular}%
}

\end{table}

\begin{table}[t]
    \centering
    \caption{\textbf{Comparison of various feature fusion methods:} We compared different fusion methods on 10 source domain and target domain pairs in the HHAR dataset. The cross-attention mechanism demonstrates superior ability to fuse the locally and globally extracted features compared to others}
    \label{tab:cross-attn}
    \begin{tabular}{c|ccc}
    \toprule
        source → target & addition & concatenation & cross attention \\
        \midrule
 0 → 2 & 0.6849 & 0.7857 & 0.8235 \\
 1 → 6 & 0.8884 & 0.9163 & 0.9163 \\
 2 → 4 & 0.4462 & 0.3386 & 0.9363 \\
 4 → 0 & 0.2620 & 0.2576 & 0.3886 \\
 4 → 1 & 0.6679 & 0.5672 & 0.9627 \\
 5 → 1 & 0.8582 & 0.8694 & 0.9851 \\
 7 → 1 & 0.8918 & 0.8545 & 0.9478 \\
 7 → 5 & 0.4402 & 0.5637 & 0.8147 \\
 8 → 3 & 0.8996 & 0.9738 & 0.9738 \\
 8 → 4 & 0.6175 & 0.7331 & 0.9681 \\
 \midrule
 Avg & 0.6656 & 0.6860 & 0.8717  \\
 \bottomrule
    \end{tabular}
    
\end{table}

\textbf{Ablation studies of loss function:}
To assess the effectiveness of invariant feature learning and alignment across diverse domains, we carried out an ablation study and presented the results in Table~\ref{tab:ablation}. Since all baselines except RAINCOAT use TCN as backbone, we test the UDA performance of TCN combined with our designed losses. As shown in the first column, the performance surpasses all baselines that were not specifically designed for time series (columns 1-8 in Table~\ref{tab:acc}), demonstrating that our designed loss function can effectively align the target domain with the source domain. In the second column, when only the classification loss $\mathcal{L}_{cls}$ is utilized, we term it a 'source-only' model, trained solely on the source domain. Despite lacking any alignment, our model demonstrates comparable performance to many baselines. Upon comparing the third and last columns, the incorporation of adversarial learning strategies leads to a significant performance enhancement in the target domain. This suggests that adversarial learning empowers the model to capture domain-invariant features. Additionally, the center loss $\mathcal{L}_{center}$ yields an absolute improvement across all datasets, as evidenced in the fourth column. This indicates that the center loss effectively aligns the features of the same class between the target domain and the source domain, demonstrating its effectiveness and necessity. To examine the impact of the margin loss $\mathcal{L}_{margin}$ and the DTW loss $\mathcal{L}_{dtw}$, we individually evaluated the prediction results by removing each one (the $5^{\text{th}}$ and $6^{\text{th}}$ columns in Table~\ref{tab:ablation}). The results demonstrate a varying degree of decrease in accuracy compared to the entire framework (the last column), indicating that both the margin and DTW loss are indispensable, playing crucial roles in the entire framework. Specifically, without the margin loss, the model-learned features may not be sufficiently aligned in the feature space, thereby impacting the effectiveness of the center loss. Similarly, without the DTW loss, the model fails to adequately learn features robust to time-step shifts, resulting in decreased accuracy.

\begin{figure*}[t]
\centering  
\subfigure[HHAR 8 - HHAR 3]{  
\begin{minipage}[t]{0.29\linewidth}
\centering    
\includegraphics[width=\linewidth]{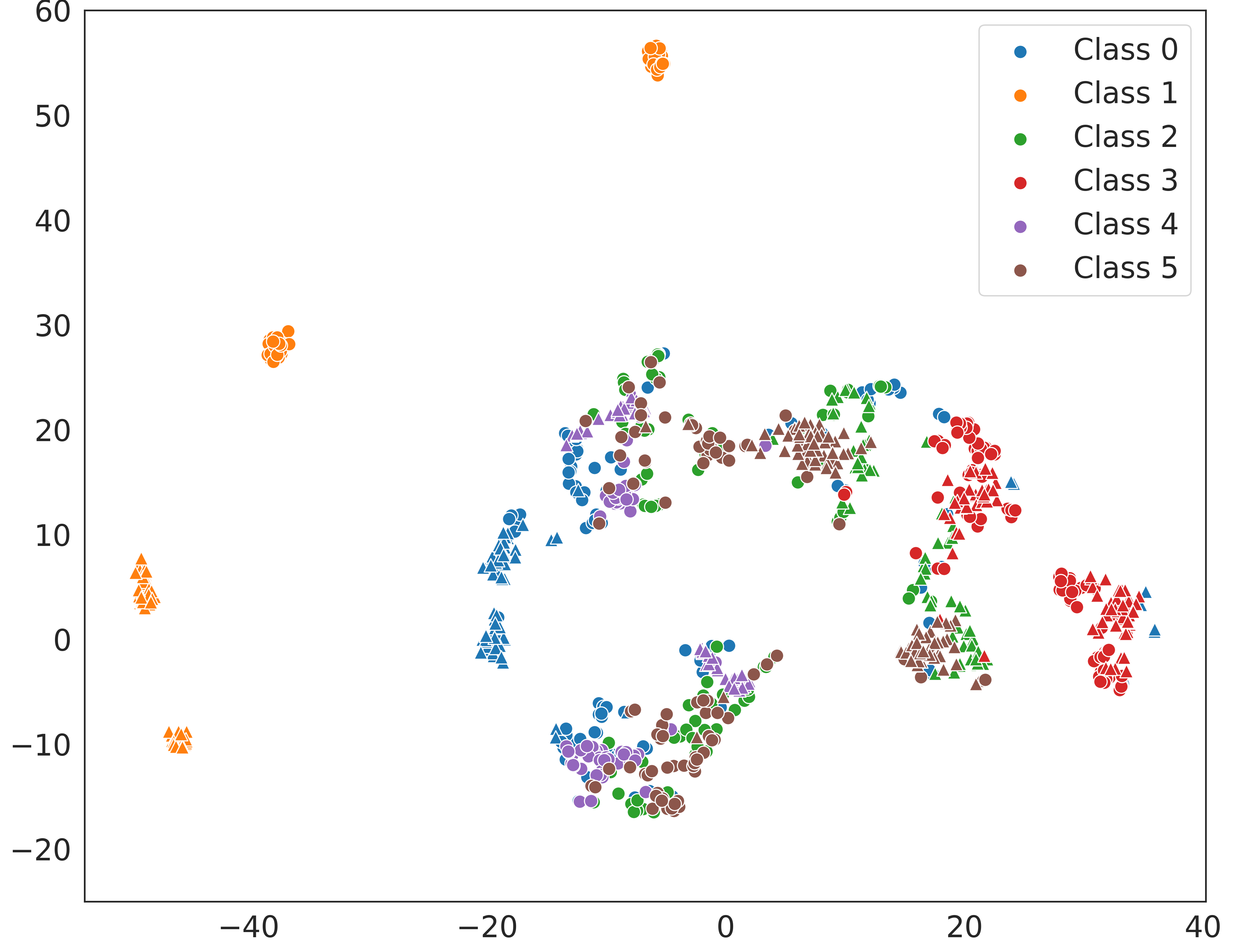} 
\end{minipage}
}
\subfigure[\abbr w/o alignment]{ 
\label{fig:tsne_cls}
\begin{minipage}[t]{0.29\linewidth}
\centering    
\includegraphics[width=\linewidth]{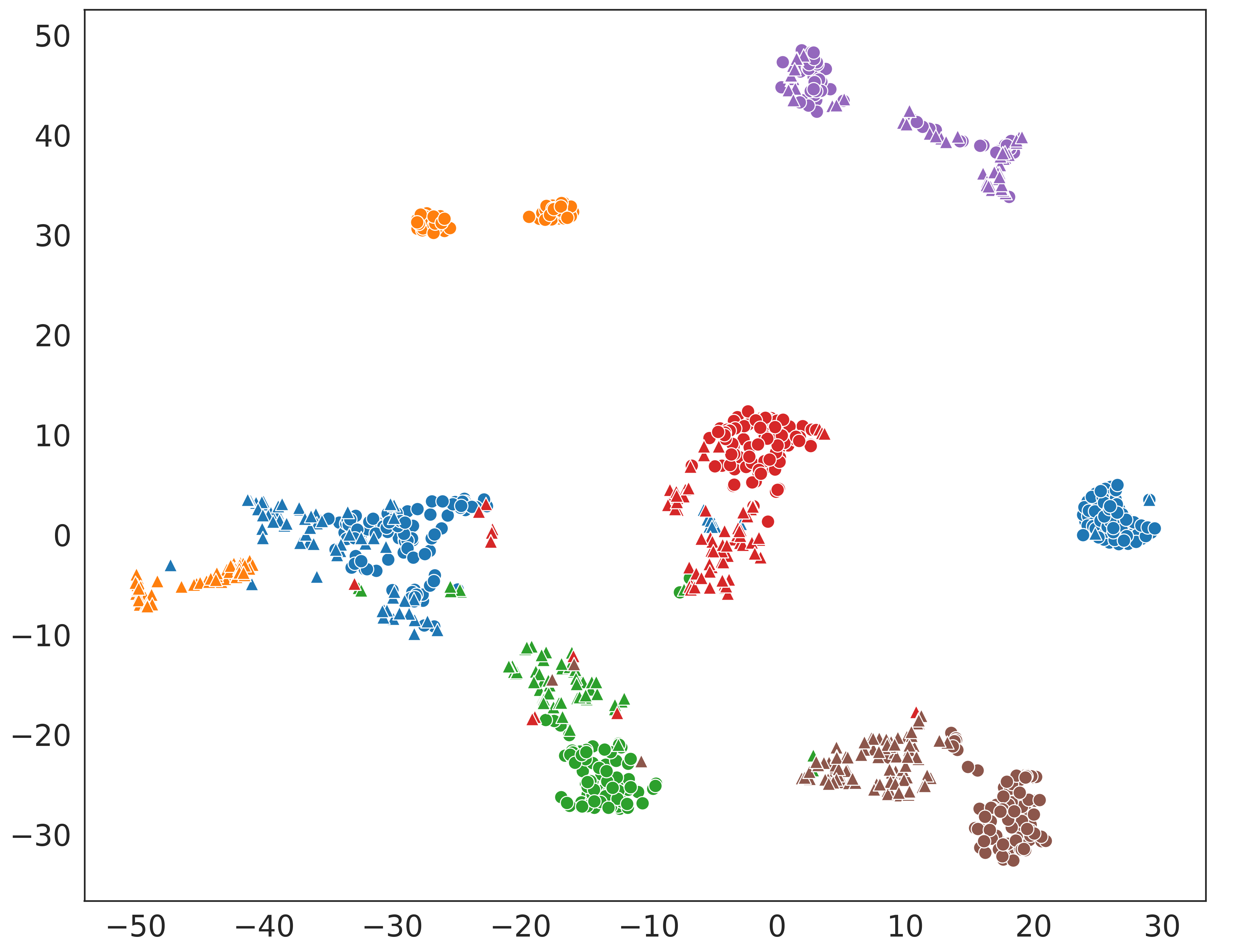} 
\end{minipage}
}
\subfigure[\abbr]{ 
\begin{minipage}[t]{0.29\linewidth}
\centering   
\includegraphics[width=\linewidth]{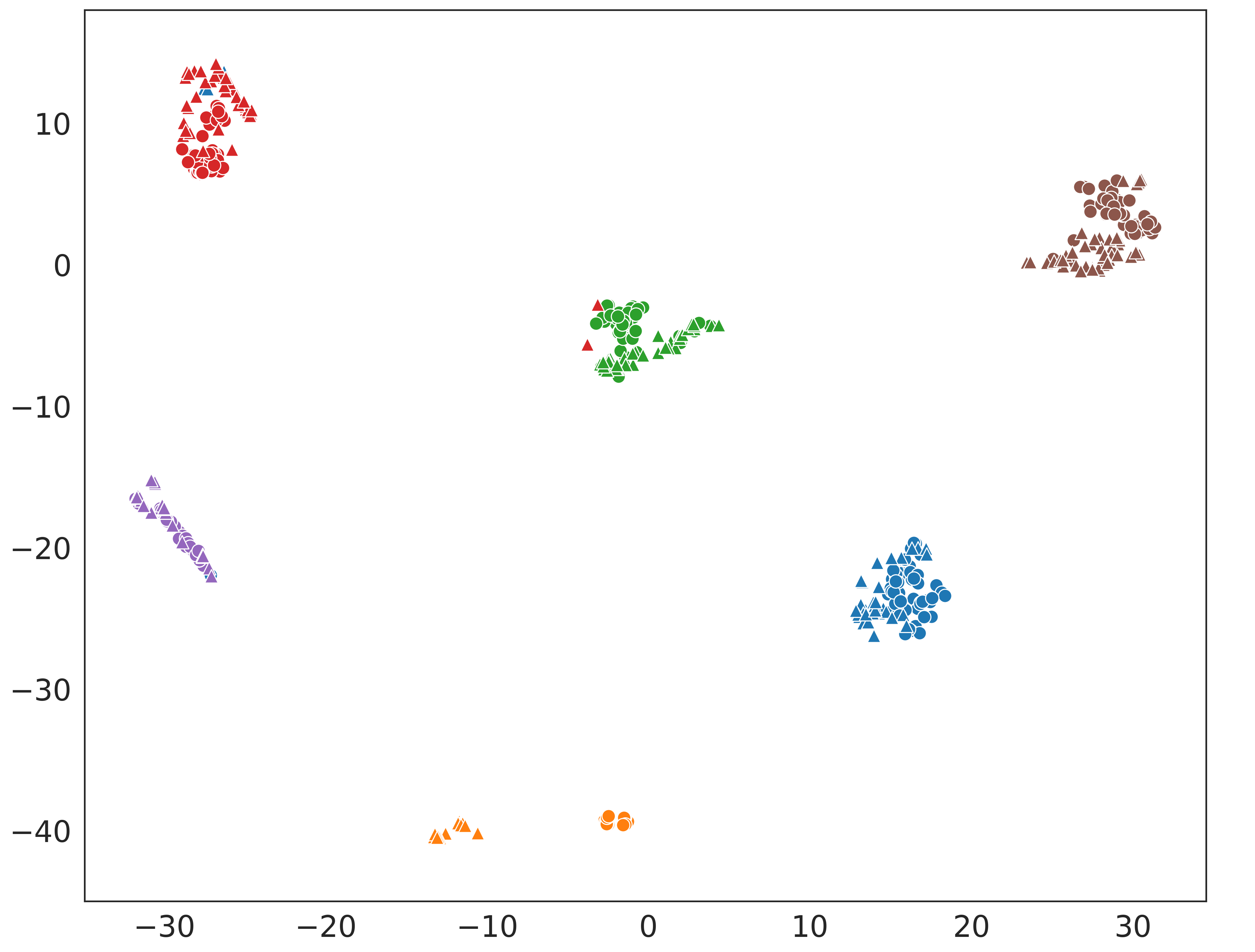}
\end{minipage}
}
\subfigure[CLUDA]{ 
\begin{minipage}[t]{0.29\linewidth}
\centering   
\includegraphics[width=\linewidth]{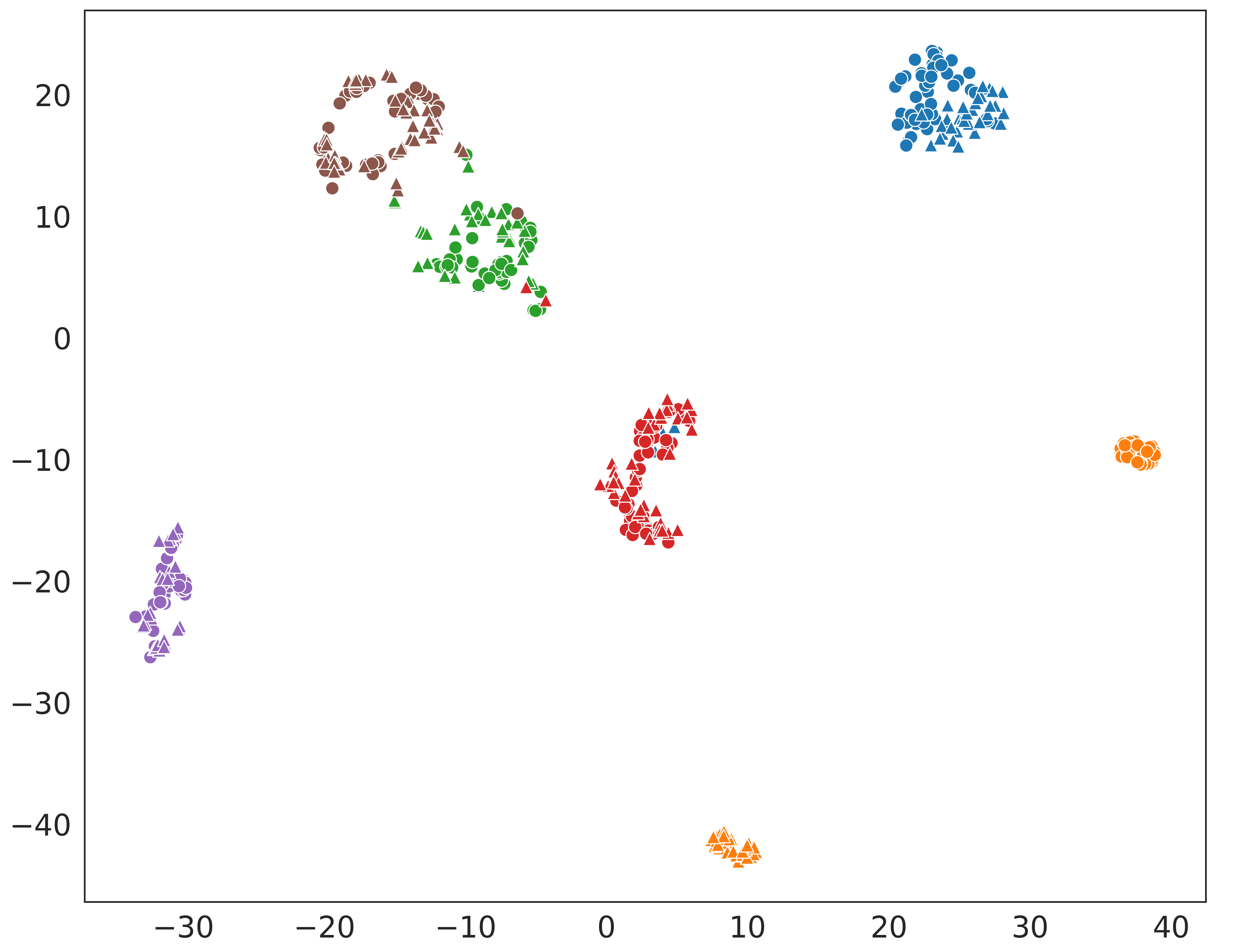}
\end{minipage}
}
\subfigure[RAINCOAT]{ 
\begin{minipage}[t]{0.29\linewidth}
\centering   
\includegraphics[width=\linewidth]{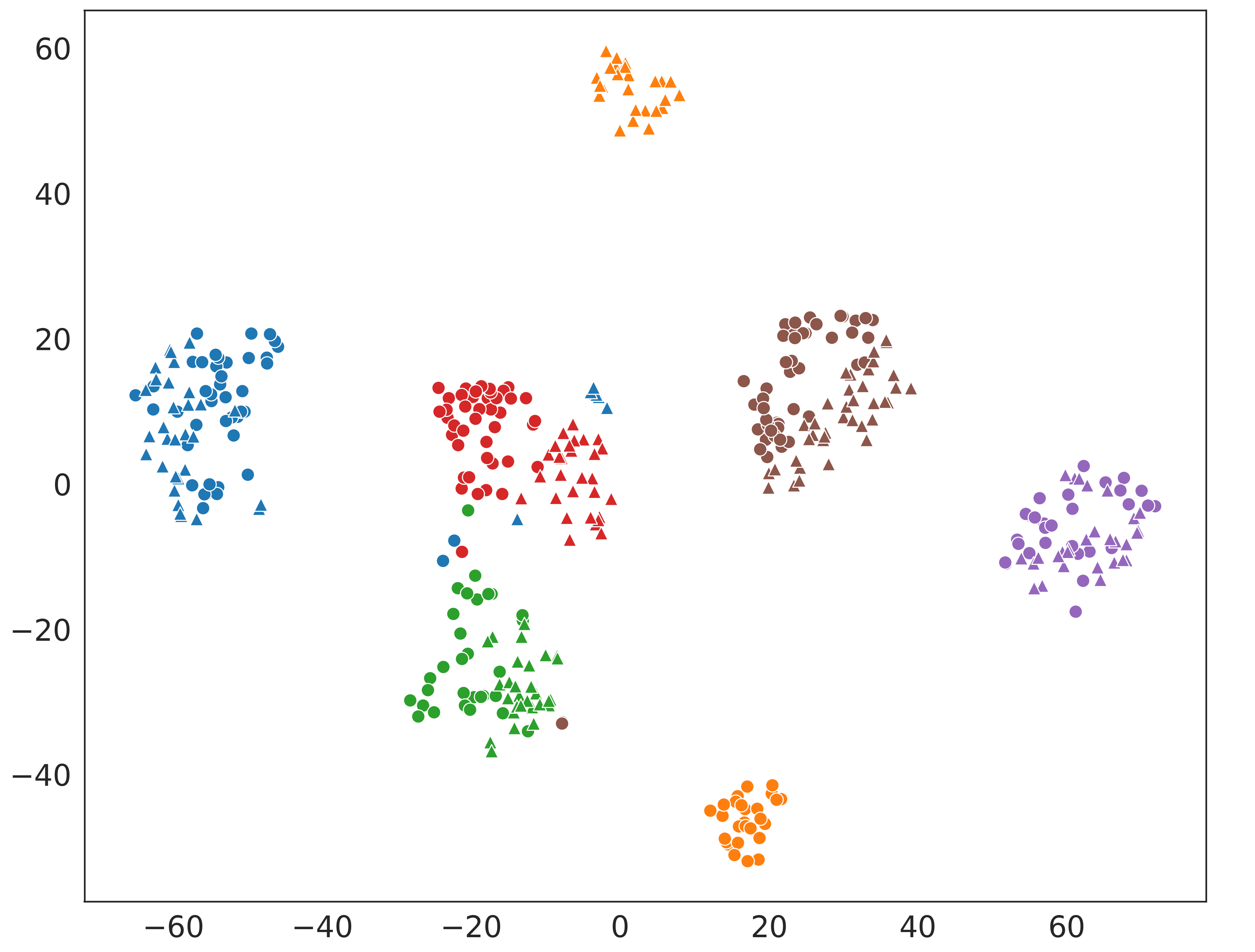}
\end{minipage}
}
\centering
\vspace{-0.2cm}
\caption{\textbf{T-SNE plots of different methods and raw data on the HHAR dataset (Domain 8 → 3)}. Each color corresponds to a different class. The circle markers represent source samples, while the triangle markers represent target samples. These T-SNE plots demonstrate that \abbr concentrates features from the same category more effectively and aligns the target domain with the source domain more accurately. }
\vspace{-0.2cm}
\label{fig:tsne_app}
\end{figure*}



\textbf{Ablation studies of model architecture:}
In order to explore different model architectures and investigate their generalization capabilities, we conduct a series of ablation experiments on the source domain. All models are trained only on source domain data using the classification loss $\mathcal{L}_{cls}$, and ultimately tested on the target domain to isolate their performance from other variables. We perform the ablation study using HHAR dataset and present results in Table~\ref{tab:ab_model}. We initially attempt the commonly used backbone, TCN~\cite{bai2018empirical}, in time series analysis ($1^{\text{st}}$ row in Table~\ref{tab:ab_model}). It was evident that TCN exhibits significantly lower generalization capability compared to other backbones. We believe this might be attributed to TCN's dilated convolution mechanism, which may not effectively extract information from continuous time series data. Next, we attempt using vanilla Transformer and PatchTST~\cite{Yuqietal-2023-PatchTST} as backbones ($2^{\text{nd}}$ and $3^{\text{st}}$ row in Table~\ref{tab:ab_model}). Both show significant improvements compared to TCN, indicating the superior effectiveness of the Transformer architecture. Notably, PatchTST outperforms the standard Transformer, highlighting the advantage of patching to direct the attention of the Transformer toward local information. Subsequently, we experiment with the two-branch architecture ($4^{\text{th}}$ to $7^{\text{th}}$ row in Table~\ref{tab:ab_model}), where each branch specializes in extracting either global or local features. When employing our global encoder and local encoder ($6^{\text{th}}$ row in Table~\ref{tab:ab_model}), the performance surpasses that of the previous models. Furthermore, with the adoption of the multi-scale local encoder, which corresponds to our \abbr, the performance is the best, demonstrating a noticeable enhancement in generalization.

\textbf{Ablation studies of fusion methods:} We compared cross-attention with other feature fusion methods such as concatenation and addition, and the experimental results are shown in the Tab.~\ref{tab:cross-attn}. We tested our model on 10 different source domain and target domain pairs in the HHAR dataset, using concatenation, addition, and cross-attention to fuse global and local features through our Local-Global Fusion Module. Clearly, the cross-attention mechanism demonstrates superior ability to fuse the locally and globally extracted features compared to simple addition or concatenation, resulting in the best UDA performance.

\subsection{Visualization}

\textbf{Visualizations of cross-attention from fusion module:} The cross-attention operation is able to extract useful robust context information between global and local features and is commonly used in other fields~\cite{Zhu_2022_CVPR}. For a more intuitive understanding of the impact of multi-scale operations and the fusion between global and local features, we visualize the cross-attention weights computed between local and global representation within the fusion module. Taking the HAR dataset as an example, in \figurename~\ref{fig:attn}, the middle section displays the heat map of attention weights. The horizontal and vertical axes represent local and global representation lengths. The waveform chart above illustrates the average weight at each position. Results for kernel sizes of 4, 8, and 16 are shown from left to right, representing different scale local encoders. The figure below represents the original time series data. The locations with higher attention weights are highlighted. 

Thanks to the modeling capability of Transformers for long dependencies, we extract the global features of the data and force the global features to be time-shift invariant through DTW loss. And the local features extracted by the convolutional network remain in temporal order, which can be matched with the patch-wise global feature. The global feature summarizes the content of a sequence, often leading to a compact representation. Local features, on the other hand, comprise pattern information about specific regions, which are especially useful for classification. As shown in the heatmap in \figurename~\ref{fig:attn}, higher similarity represents more useful information for classification. This observation indicates that local features with more relevant information to global features are more important. Therefore, based on similarity score, further selecting local features can yield features that contain more useful contextual information. Besides, thanks to the multi-scale operation, \abbr is able to thoroughly attend to sufficiently detailed local features, which not only aids in learning meaningful features but also facilitates the alignment between features. Therefore, our proposed method allows the extraction of category-specific features, achieving effective feature fusion conducive to classification.

\textbf{T-SNE visualizations of learned representations:}We generated T-SNE plots of learned embeddings for different methods. In \figurename~\ref{fig:tsne_app}, we present the T-SNE plots of the original data, RAINCOAT~\cite{he2023domain}, CLUDA~\cite{ozyurt2023contrastive}, \abbr without any alignment and the complete \abbr for HHAR dataset of adapting source 8 to target 3. Despite the original data being challenging to discern, \abbr is still capable of effectively distinguishing between different categories and aligning the features of the target domain with the source domain. When no alignment is applied, $i.e.$ the model is only trained on source domain, the domain gap between source domain and target domain is obvious. Compared to RAINCOAT and CLUDA, our clustering results are also more concentrated. Specifically, for Class 1, which experiences significant domain shift, \abbr achieves a much better alignment than CLUDA and RAINCOAT. This suggests that \abbr effectively adapts the model to the target domain, leading to improved performance and more accurate predictions. These findings also demonstrate the efficacy of \abbr for domain adaptation and highlight its potential for a wide range of applications, including robotics, healthcare, and sports performance analysis. 

\subsection{Inference Time and Parameters}\label{sec_app:params}

\begin{figure}[t]
\centering  
\subfigure[\textbf{}]{  
\begin{minipage}[t]{0.46\columnwidth}
\centering    
\includegraphics[width=1.1\columnwidth]{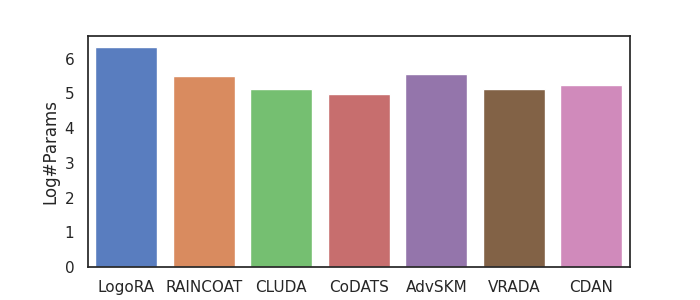} 
    \label{fig:para}
\end{minipage}
}
\subfigure[\textbf{}]{ 
\begin{minipage}[t]{0.46\columnwidth}
\centering   
\includegraphics[width=1.1\columnwidth]{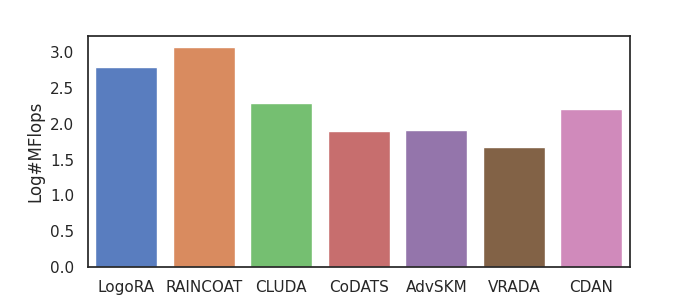}
    \label{fig:flops}
\end{minipage}
}
\centering
\vspace{-0.2cm}
\caption{\textbf{(a):} Logarithmically transformed parameter count of different methods \textbf{(b):} Logarithmically transformed million floating point operations per second of different methods}
\label{fig:inf}
\end{figure}

We computed and compared the computational complexity and parameter count of \abbr with other time series DA methods, as well as the general DA method CDAN. As illustrated in the \figurename~\ref{fig:inf}, our \abbr incurs only a slight increase in computational complexity and the size of model parameters, but leads to a significant improvement in generalization (Sec.~\ref{sec:exp}).

\begin{figure*}[t]
\begin{center}
\includegraphics[width=0.925\linewidth]{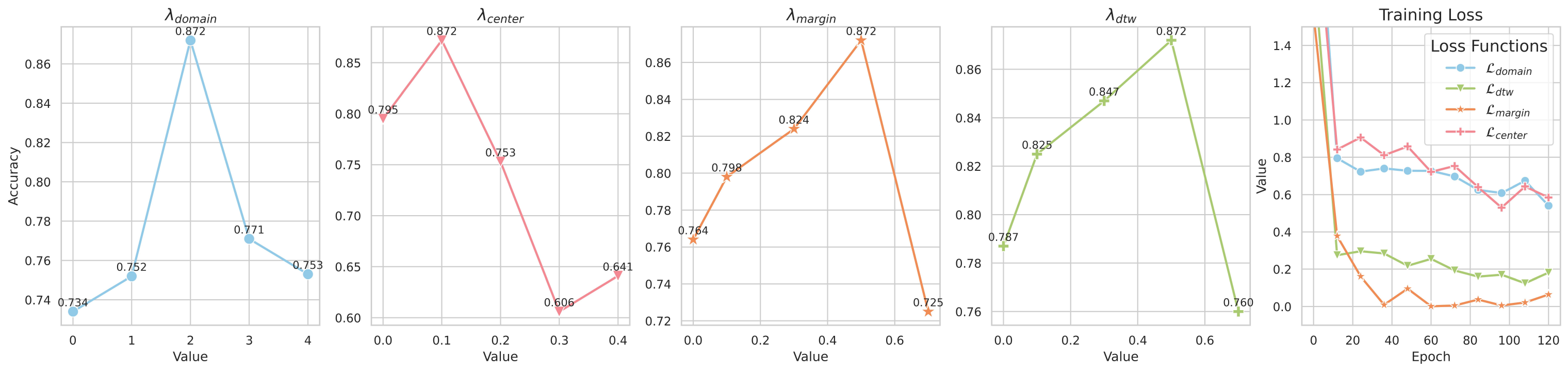}
\end{center}
\caption{\textbf{Sensitivity Analysis:} We perform a comprehensive sensitivity analysis on the four key hyperparameters of each loss functions and present the results on HHAR dataset in the first four plots. The last image displays the loss curves, illustrating the training stability of \abbr.}
\label{fig:sen}
\end{figure*}

\subsection{Symbols and Notations}
We present the symbols and notations in Tab.~\ref{tab:my_label}
\begin{table}[t]
    \centering
    \caption{\textbf{Symbols and Notations}}
    \begin{tabular}{ll}
    \toprule
    Symbols & Notations \\ \midrule
        $x^s$$\quad\quad\quad\quad\quad\quad\;$ & input from source domain \\
        $x^t$$\quad\quad\quad\quad\quad\quad\;$ & input from target domain \\

        $y^s$$\quad\quad\quad\quad\quad\quad\;$ & label from source domain \\
        $y^t$$\quad\quad\quad\quad\quad\quad\;$ & label from target domain \\

        $\mathcal{D}_S\quad\quad\quad\quad\quad\quad\;$ & source domain\\

        $\mathcal{D}_T\quad\quad\quad\quad\quad\quad\;$ & target domain\\

        $\mathcal{S}\quad\quad\quad\quad\quad\quad\;$ & source domain dataset\\

        $\mathcal{T}\quad\quad\quad\quad\quad\quad\;$ & target domain dataset\\

        $o\quad\quad\quad\quad\quad\quad\;$ & patch\\

        $z_g\quad\quad\quad\quad\quad\quad\;$ & global representation\\

        $z_l\quad\quad\quad\quad\quad\quad\;$ & local representation\\

        $\hat{z}\quad\quad\quad\quad\quad\quad\;$ & fused representation\\
        $c_j\quad\quad\quad\quad\quad\quad\;$ & the $j$th class prototype\\

        $F(\cdot)\quad\quad\quad\quad\quad\quad\;$ & feature extractor\\
        $G(\cdot)\quad\quad\quad\quad\quad\quad\;$ & fusion module\\
        $C(\cdot)\quad\quad\quad\quad\quad\quad\;$ & classifier\\
        $D(\cdot)\quad\quad\quad\quad\quad\quad\;$ & domain discriminator\\
        $T\quad\quad\quad\quad\quad\quad\;$ & number of time steps\\
        $N\quad\quad\quad\quad\quad\quad\;$ & number of different scale local encoders\\
        $M\quad\quad\quad\quad\quad\quad\;$ & number of patches\\
        $P\quad\quad\quad\quad\quad\quad\;$ & length of patch\\

        $S\quad\quad\quad\quad\quad\quad\;$ & stride between adjacent patches \\ \bottomrule

    \end{tabular}
    
    \label{tab:my_label}
\end{table}

\subsection{Sensitivity and Training Stability Analysis}\label{sec:sensitivity}

We perform a comprehensive sensitivity analysis on the four key hyperparameters of each loss functions. The analysis involves varying each hyperparameter while keeping the others fixed, to observe their individual impact on the model's performance. We present the results in terms of accuracy in the first four plots of \figurename~\ref{fig:sen}. Our experiments reveal that all the designed loss functions have critical impact on the final UDA performance. Besides, the optimal performance is achieved at specific configurations, which also demonstrates that these loss functions work well together.

We also assess the training stability by monitoring the loss and provide the visualization in the last image of \figurename~\ref{fig:sen}. The training process demonstrated stability, as indicated by the consistent convergence of all loss functions. The stable training process also reveals that our designed loss functions complement one another effectively, resulting in the successful alignment of diverse features.

\section{Conclusion and Limitations}

Through an investigation of previous works on UDA for time series, we find that existing methods do not sufficiently explore global and local features as well as the alignments in time series, thereby limiting the model's generalization capability. To this end, we propose \abbr, a novel framework that takes advantage of both global and local representations of time series data. Furthermore, we devise loss functions based on DTW and triplet loss to learn time-shift invariant features in the source domain. We then employ adversarial training and metric-based methods to further align features across different domains. As a result, \abbr achieves the state-of-the-art performance. Extensive ablation studies demonstrate the fusion of global and local representations and the alignment losses both yield clear performance improvements. The visualization results intuitively demonstrate the impact of multi-scale operations and the superior generalization capability of \abbr.

We also acknowledge that the introduction of new model architectures leads to an increase in the number of parameters. We believe that there is potential for new, lighter-weight models to be developed for extracting global and local features from time series. Additionally, while we designed some alignment loss functions, there is room for the development of more efficient adaptation methods. Moreover, time series data are often coupled with static data such as images or text. Therefore, encoding inputs from various modalities, such as Multi-modal Large Language Models (MLLM)~\cite{jin2024position}, could be a promising expansion of \abbr.


\bibliographystyle{IEEEtran}
\bibliography{IEEEabrv, logora}

\begin{thebibliography}{10}
\providecommand{\url}[1]{#1}
\csname url@samestyle\endcsname
\providecommand{\newblock}{\relax}
\providecommand{\bibinfo}[2]{#2}
\providecommand{\BIBentrySTDinterwordspacing}{\spaceskip=0pt\relax}
\providecommand{\BIBentryALTinterwordstretchfactor}{4}
\providecommand{\BIBentryALTinterwordspacing}{\spaceskip=\fontdimen2\font plus
\BIBentryALTinterwordstretchfactor\fontdimen3\font minus \fontdimen4\font\relax}
\providecommand{\BIBforeignlanguage}[2]{{%
\expandafter\ifx\csname l@#1\endcsname\relax
\typeout{** WARNING: IEEEtran.bst: No hyphenation pattern has been}%
\typeout{** loaded for the language `#1'. Using the pattern for}%
\typeout{** the default language instead.}%
\else
\language=\csname l@#1\endcsname
\fi
#2}}
\providecommand{\BIBdecl}{\relax}
\BIBdecl

\bibitem{koh2021wilds}
P.~W. Koh, S.~Sagawa, H.~Marklund, S.~M. Xie, M.~Zhang, A.~Balsubramani, W.~Hu, M.~Yasunaga, R.~L. Phillips, I.~Gao \emph{et~al.}, ``Wilds: A benchmark of in-the-wild distribution shifts,'' in \emph{ICML}, 2021.

\bibitem{wen2022robust}
Q.~Wen, L.~Yang, T.~Zhou, and L.~Sun, ``Robust time series analysis and applications: An industrial perspective,'' in \emph{SIGKDD}, 2022.

\bibitem{tang2019tensor}
Y.~Tang, Y.~Xie, X.~Yang, J.~Niu, and W.~Zhang, ``Tensor multi-elastic kernel self-paced learning for time series clustering,'' \emph{TKDE}, 2019.

\bibitem{borges2022iot}
J.~B. Borges, J.~P. Medeiros, L.~P. Barbosa, H.~S. Ramos, and A.~A. Loureiro, ``Iot botnet detection based on anomalies of multiscale time series dynamics,'' \emph{TKDE}, 2022.

\bibitem{wu2021early}
R.~Wu, A.~Der, and E.~J. Keogh, ``When is early classification of time series meaningful?'' \emph{TKDE}, 2021.

\bibitem{ravuri2021skilful}
S.~Ravuri, K.~Lenc, M.~Willson, D.~Kangin, R.~Lam, P.~Mirowski, M.~Fitzsimons, M.~Athanassiadou, S.~Kashem, S.~Madge \emph{et~al.}, ``Skilful precipitation nowcasting using deep generative models of radar,'' \emph{Nature}, 2021.

\bibitem{jin2022multivariate}
M.~Jin, Y.~Zheng, Y.-F. Li, S.~Chen, B.~Yang, and S.~Pan, ``Multivariate time series forecasting with dynamic graph neural odes,'' \emph{TKDE}, 2022.

\bibitem{li2020efficient}
G.~Li, B.~Choi, J.~Xu, S.~S. Bhowmick, K.-P. Chun, and G.~L.-H. Wong, ``Efficient shapelet discovery for time series classification,'' \emph{TKDE}, 2020.

\bibitem{purushotham2016variational}
S.~Purushotham, W.~Carvalho, T.~Nilanon, and Y.~Liu, ``Variational recurrent adversarial deep domain adaptation,'' in \emph{ICLR}, 2016.

\bibitem{ozyurt2023contrastive}
Y.~Ozyurt, S.~Feuerriegel, and C.~Zhang, ``Contrastive learning for unsupervised domain adaptation of time series,'' \emph{ICLR}, 2023.

\bibitem{cai2021time}
R.~Cai, J.~Chen, Z.~Li, W.~Chen, K.~Zhang, J.~Ye, Z.~Li, X.~Yang, and Z.~Zhang, ``Time series domain adaptation via sparse associative structure alignment,'' in \emph{AAAI}, 2021.

\bibitem{liu2021adversarial}
Q.~Liu and H.~Xue, ``Adversarial spectral kernel matching for unsupervised time series domain adaptation.'' in \emph{IJCAI}, 2021.

\bibitem{wilson2020multi}
G.~Wilson, J.~R. Doppa, and D.~J. Cook, ``Multi-source deep domain adaptation with weak supervision for time-series sensor data,'' in \emph{Proceedings of the 26th ACM SIGKDD international conference on knowledge discovery \& data mining}, 2020.

\bibitem{he2023domain}
H.~He, O.~Queen, T.~Koker, C.~Cuevas, T.~Tsiligkaridis, and M.~Zitnik, ``Domain adaptation for time series under feature and label shifts,'' in \emph{ICML}, 2023.

\bibitem{yue2022ts2vec}
Z.~Yue, Y.~Wang, J.~Duan, T.~Yang, C.~Huang, Y.~Tong, and B.~Xu, ``Ts2vec: Towards universal representation of time series,'' in \emph{AAAI}, 2022.

\bibitem{bai2018empirical}
S.~Bai, J.~Z. Kolter, and V.~Koltun, ``An empirical evaluation of generic convolutional and recurrent networks for sequence modeling,'' \emph{arXiv preprint arXiv:1803.01271}, 2018.

\bibitem{cui2016multi}
Z.~Cui, W.~Chen, and Y.~Chen, ``Multi-scale convolutional neural networks for time series classification,'' \emph{arXiv preprint arXiv:1603.06995}, 2016.

\bibitem{zhou2021informer}
H.~Zhou, S.~Zhang, J.~Peng, S.~Zhang, J.~Li, H.~Xiong, and W.~Zhang, ``Informer: Beyond efficient transformer for long sequence time-series forecasting,'' in \emph{AAAI}, 2021.

\bibitem{muller2007dynamic}
M.~M{\"u}ller, ``Dynamic time warping,'' \emph{Information retrieval for music and motion}, 2007.

\bibitem{ben2010theory}
S.~Ben-David, J.~Blitzer, K.~Crammer, A.~Kulesza, F.~Pereira, and J.~W. Vaughan, ``A theory of learning from different domains,'' \emph{Machine learning}, 2010.

\bibitem{ganin2016domain}
Y.~Ganin, E.~Ustinova, H.~Ajakan, P.~Germain, H.~Larochelle, F.~Laviolette, M.~Marchand, and V.~Lempitsky, ``Domain-adversarial training of neural networks,'' \emph{The journal of machine learning research}, 2016.

\bibitem{long2018conditional}
M.~Long, Z.~Cao, J.~Wang, and M.~I. Jordan, ``Conditional adversarial domain adaptation,'' \emph{NeurIPS}, 2018.

\bibitem{tzeng2017adversarial}
E.~Tzeng, J.~Hoffman, K.~Saenko, and T.~Darrell, ``Adversarial discriminative domain adaptation,'' in \emph{CVPR}, 2017.

\bibitem{xu2020adversarial}
M.~Xu, J.~Zhang, B.~Ni, T.~Li, C.~Wang, Q.~Tian, and W.~Zhang, ``Adversarial domain adaptation with domain mixup,'' in \emph{AAAI}, 2020.

\bibitem{pei2018multi}
Z.~Pei, Z.~Cao, M.~Long, and J.~Wang, ``Multi-adversarial domain adaptation,'' in \emph{AAAI}, 2018.

\bibitem{zhang2023free}
Y.~Zhang, X.~Wang, J.~Liang, Z.~Zhang, L.~Wang, R.~Jin, and T.~Tan, ``Free lunch for domain adversarial training: Environment label smoothing,'' \emph{ICLR}, 2023.

\bibitem{rozantsev2018beyond}
A.~Rozantsev, M.~Salzmann, and P.~Fua, ``Beyond sharing weights for deep domain adaptation,'' \emph{TPAMI}, 2018.

\bibitem{sun2016deep}
B.~Sun and K.~Saenko, ``Deep coral: Correlation alignment for deep domain adaptation,'' in \emph{ECCV Workshops}, 2016.

\bibitem{zhu2020deep}
Y.~Zhu, F.~Zhuang, J.~Wang, G.~Ke, J.~Chen, J.~Bian, H.~Xiong, and Q.~He, ``Deep subdomain adaptation network for image classification,'' \emph{TNNLS}, 2020.

\bibitem{chen2020homm}
C.~Chen, Z.~Fu, Z.~Chen, S.~Jin, Z.~Cheng, X.~Jin, and X.-S. Hua, ``Homm: Higher-order moment matching for unsupervised domain adaptation,'' in \emph{AAAI}, 2020.

\bibitem{rahman2020minimum}
M.~M. Rahman, C.~Fookes, M.~Baktashmotlagh, and S.~Sridharan, ``On minimum discrepancy estimation for deep domain adaptation,'' \emph{Domain Adaptation for Visual Understanding}, 2020.

\bibitem{kang2019contrastive}
G.~Kang, L.~Jiang, Y.~Yang, and A.~G. Hauptmann, ``Contrastive adaptation network for unsupervised domain adaptation,'' in \emph{CVPR}, 2019.

\bibitem{singh2021clda}
A.~Singh, ``Clda: Contrastive learning for semi-supervised domain adaptation,'' \emph{NeurIPS}, 2021.

\bibitem{huang2021model}
J.~Huang, D.~Guan, A.~Xiao, and S.~Lu, ``Model adaptation: Historical contrastive learning for unsupervised domain adaptation without source data,'' \emph{NeurIPS}, 2021.

\bibitem{tang2021gradient}
S.~Tang, P.~Su, D.~Chen, and W.~Ouyang, ``Gradient regularized contrastive learning for continual domain adaptation,'' in \emph{AAAI}, 2021.

\bibitem{cuturi2016smoothed}
M.~Cuturi and G.~Peyr{\'e}, ``A smoothed dual approach for variational wasserstein problems,'' \emph{SIAM Journal on Imaging Sciences}, 2016.

\bibitem{jin2022domain}
X.~Jin, Y.~Park, D.~Maddix, H.~Wang, and Y.~Wang, ``Domain adaptation for time series forecasting via attention sharing,'' in \emph{ICML}, 2022.

\bibitem{tsymbal2004problem}
A.~Tsymbal, ``The problem of concept drift: definitions and related work,'' \emph{Computer Science Department, Trinity College Dublin}, vol. 106, 2004.

\bibitem{wen2024onenet}
Q.~Wen, W.~Chen, L.~Sun, Z.~Zhang, L.~Wang, R.~Jin, T.~Tan \emph{et~al.}, ``Onenet: Enhancing time series forecasting models under concept drift by online ensembling,'' \emph{NeurIPS}, 2024.

\bibitem{anderson2004towards}
T.~Anderson, ``Towards a theory of online learning,'' \emph{Theory and practice of online learning}, 2004.

\bibitem{zhang2023domain}
Y.-F. Zhang, J.~Wang, J.~Liang, Z.~Zhang, B.~Yu, L.~Wang, D.~Tao, and X.~Xie, ``Domain-specific risk minimization for domain generalization,'' in \emph{SIGKDD}, 2023.

\bibitem{zhang2023adanpc}
Y.~Zhang, X.~Wang, K.~Jin, K.~Yuan, Z.~Zhang, L.~Wang, R.~Jin, and T.~Tan, ``Adanpc: Exploring non-parametric classifier for test-time adaptation,'' in \emph{ICML}, 2023.

\bibitem{LOGO}
Y.~Wang, M.~Wu, R.~Jin, X.~Li, L.~Xie, and Z.~Chen, ``Local-global correlation fusion-based graph neural network for remaining useful life prediction,'' \emph{TNNLS}, vol.~PP, 11 2023.

\bibitem{sen2019think}
R.~Sen, H.-F. Yu, and I.~S. Dhillon, ``Think globally, act locally: A deep neural network approach to high-dimensional time series forecasting,'' \emph{NeurIPS}, 2019.

\bibitem{Tang_joint_2020}
X.~Tang, H.~Yao, Y.~Sun, C.~Aggarwal, P.~Mitra, and S.~Wang, ``Joint modeling of local and global temporal dynamics for multivariate time series forecasting with missing values,'' \emph{AAAI}, 2020.

\bibitem{wang2023micn}
H.~Wang, J.~Peng, F.~Huang, J.~Wang, J.~Chen, and Y.~Xiao, ``Micn: Multi-scale local and global context modeling for long-term series forecasting,'' in \emph{The eleventh ICLR}, 2023.

\bibitem{Yuqietal-2023-PatchTST}
Y.~Nie, N.~H.~Nguyen, P.~Sinthong, and J.~Kalagnanam, ``A time series is worth 64 words: Long-term forecasting with transformers,'' in \emph{ICLR}, 2023.

\bibitem{kwapisz2011activity}
J.~R. Kwapisz, G.~M. Weiss, and S.~A. Moore, ``Activity recognition using cell phone accelerometers,'' \emph{SigKDD}, 2011.

\bibitem{anguita2013public}
D.~Anguita, A.~Ghio, L.~Oneto, X.~Parra, J.~L. Reyes-Ortiz \emph{et~al.}, ``A public domain dataset for human activity recognition using smartphones.'' in \emph{Esann}, 2013.

\bibitem{stisen2015smart}
A.~Stisen, H.~Blunck, S.~Bhattacharya, T.~S. Prentow, M.~B. Kj{\ae}rgaard, A.~Dey, T.~Sonne, and M.~M. Jensen, ``Smart devices are different: Assessing and mitigatingmobile sensing heterogeneities for activity recognition,'' in \emph{SenSys}, 2015.

\bibitem{goldberger2000physiobank}
A.~L. Goldberger, L.~A. Amaral, L.~Glass, J.~M. Hausdorff, P.~C. Ivanov, R.~G. Mark, J.~E. Mietus, G.~B. Moody, C.-K. Peng, and H.~E. Stanley, ``Physiobank, physiotoolkit, and physionet: components of a new research resource for complex physiologic signals,'' \emph{circulation}, 2000.

\bibitem{Zhu_2022_CVPR}
H.~Zhu, W.~Ke, D.~Li, J.~Liu, L.~Tian, and Y.~Shan, ``Dual cross-attention learning for fine-grained visual categorization and object re-identification,'' in \emph{CVPR}.

\bibitem{jin2024position}
M.~Jin, Y.~Zhang, W.~Chen, K.~Zhang, Y.~Liang, B.~Yang, J.~Wang, S.~Pan, and Q.~Wen, ``Position paper: What can large language models tell us about time series analysis,'' \emph{ICML}, 2024.

\end{thebibliography}


\section{Biography Section}
 



\begin{IEEEbiography}[{\includegraphics[width=1in,height=1.25in,keepaspectratio]{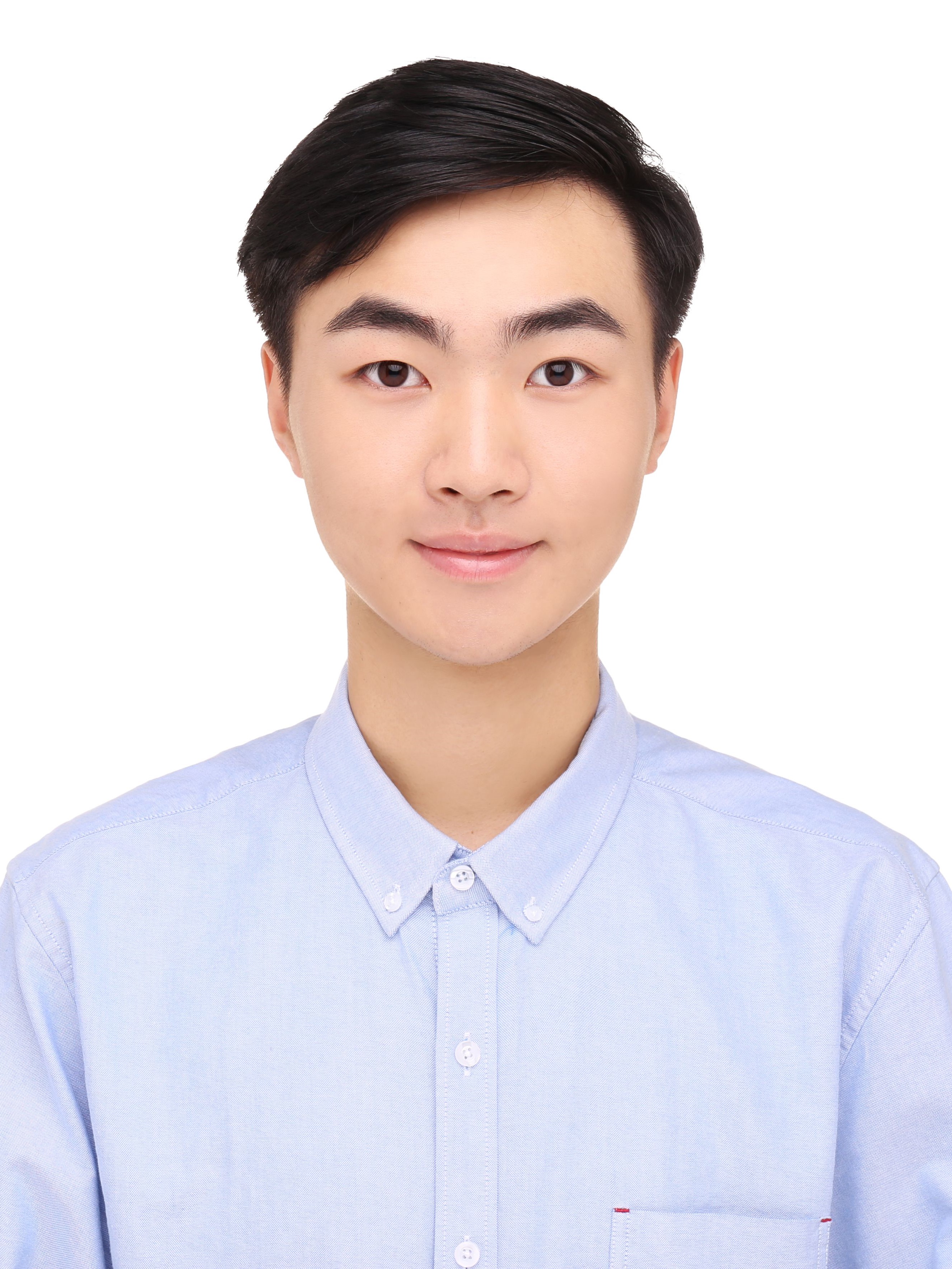}}]{Huanyu Zhang} is currently pursuing his Ph.D. degree of Computer Science at the New Laboratory of Pattern Recognition (NLPR), State Key Laboratory of Multimodal Artificial Intelligence Systems (MAIS), Institute of Automation, Chinese Academy of Sciences (CASIA). His current research interests mainly include time series analysis.
\end{IEEEbiography}

\begin{IEEEbiography}[{\includegraphics[width=1in,height=1.25in,keepaspectratio]{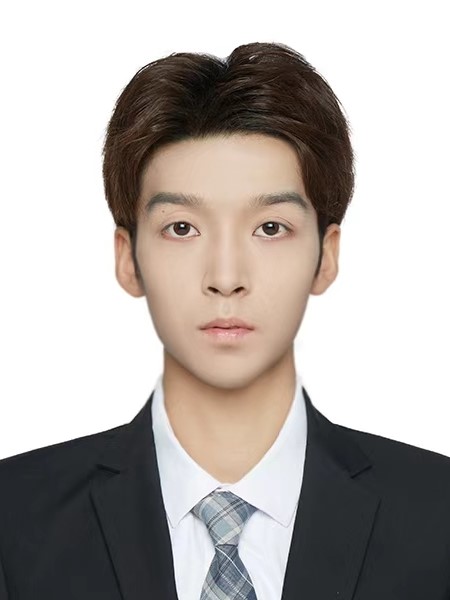}}]{Yifan Zhang} is currently pursuing his Ph.D. degree of Computer Science at the New Laboratory of Pattern Recognition (NLPR), State Key Laboratory of Multimodal Artificial Intelligence Systems (MAIS), Institute of Automation, Chinese Academy of Sciences (CASIA). His current research interests mainly include robust and reliable machine learning (ML) systems.
\end{IEEEbiography}

\begin{IEEEbiography}[{\includegraphics[width=1in,height=1.25in,keepaspectratio]{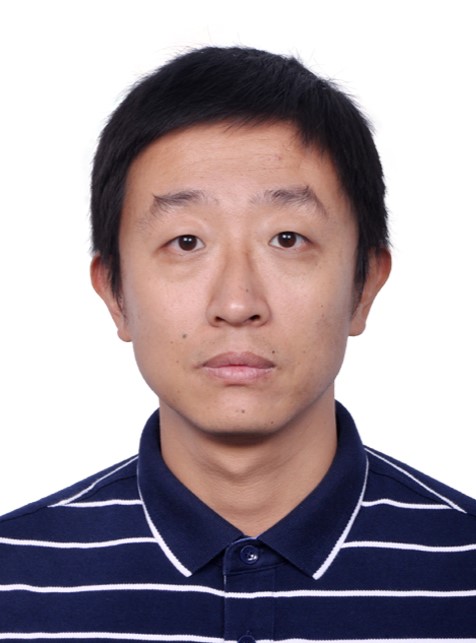}}]{Zhang Zhang} received the PhD degree from National Laboratory of Pattern Recognition (NLPR), Institute of Automation, Chinese Academy of Sciences (CASIA) in 2009. From 2009 to 2010, he was a research fellow at the School of Computer Science and Engineering, Nanyang Technological University (NTU). In September 2010, he joined the NLPR, Institute of Automation, Chinese Academy of Sciences(CASIA). Now, he is an Associate Professor at the New Laboratory of Pattern Recognition (NLPR), CASIA.
\end{IEEEbiography}

\begin{IEEEbiography}[{\includegraphics[width=1in,height=1.25in,keepaspectratio]{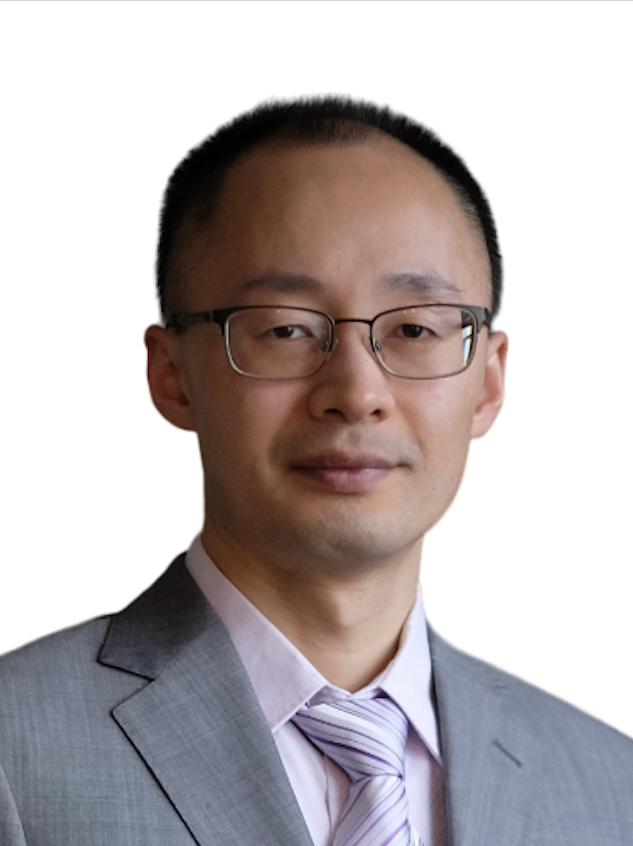}}]{Qingsong Wen} is the Head of AI Research and Chief Scientist at Squirrel Ai Learning Inc., leading a team (in both Seattle and Shanghai) working in EdTech area via AI technologies. Before that, he worked at Alibaba, Qualcomm, Marvell, etc., and received his M.S. and Ph.D. degrees in Electrical and Computer Engineering from Georgia Institute of Technology, USA. He has served as an Associate Editor for IEEE Signal Processing Letters. He is an IEEE Senior Member.
\end{IEEEbiography}

\begin{IEEEbiography}[{\includegraphics[width=1in,height=1.25in,keepaspectratio]{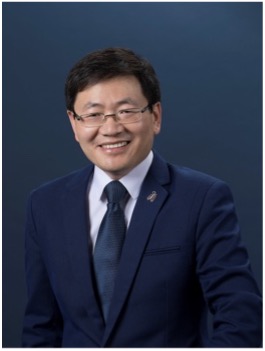}}]{Liang Wang}
received both the BEng and MEng degrees from Anhui University in 1997 and 2000, respectively, and the PhD degree from the Institute of Automation, Chinese Academy of Sciences (CASIA) in 2004. From 2004 to 2010, he was a research assistant at Imperial College London, United Kingdom, and Monash University, Australia, a research fellow at the University of Melbourne, Australia, and a lecturer at the University of Bath, United Kingdom, respectively. Currently, he is a full professor of the Hundred Talents Program at the State Key Laboratory of Multimodal Artificial Intelligence Systems, CASIA. He has widely published in highly ranked international journals such as IEEE TPAMI and IEEE TIP, and leading international conferences such as CVPR, ICCV, and ECCV. He has served as an Associate Editor of IEEE TPAMI, IEEE TIP, and PR. He is an IEEE Fellow and an IAPR Fellow.
\end{IEEEbiography}

\vfill

\end{document}